%% file: VideoStudio_camera.tex
\documentclass[runningheads]{llncs}

\usepackage{eccv}

\usepackage{eccvabbrv}

\usepackage{graphicx}
\usepackage{booktabs}

\usepackage{bm}
\usepackage{mathtools}
\usepackage{tabularx, multirow, bbding, makecell}
\newcolumntype{C}{>{\centering\arraybackslash}X}

\usepackage{lipsum}
\usepackage{graphics}
\usepackage{caption}

\usepackage[accsupp]{axessibility}  

\usepackage{hyperref}

\usepackage{orcidlink}

\begin{document}

\title{VideoStudio: Generating Consistent-Content and Multi-Scene Videos}

\titlerunning{VideoStudio}

\author{Fuchen Long \and
Zhaofan Qiu \and
Ting Yao \and
Tao Mei}


\authorrunning{Fuchen Long, Zhaofan Qiu, Ting Yao and Tao Mei}

\institute{HiDream.ai Inc. \\
\email{\{longfuchen, qiuzhaofan, tiyao, tmei\}@hidream.ai}
}

\maketitle

\begin{abstract}
	The recent innovations and breakthroughs in diffusion models have significantly expanded the possibilities of generating high-quality videos for the given prompts. Most existing works tackle the single-scene scenario with only one video event occurring in a single background. Extending to generate multi-scene videos nevertheless is not trivial and necessitates to nicely manage the logic in between while preserving the consistent visual appearance of key content across video scenes. In this paper, we propose a novel framework, namely VideoStudio, for consistent-content and multi-scene video generation. Technically, VideoStudio leverages Large Language Models (LLM) to convert the input prompt into comprehensive multi-scene script that benefits from the logical knowledge learnt by LLM. The script for each scene includes a prompt describing the event, the foreground/background entities, as well as camera movement. VideoStudio identifies the common entities throughout the script and asks LLM to detail each entity. The resultant entity description is then fed into a text-to-image model to generate a reference image for each entity. Finally, VideoStudio outputs a multi-scene video by generating each scene video via a diffusion process that takes the reference images, the descriptive prompt of the event and camera movement into account. The diffusion model incorporates the reference images as the condition and alignment to strengthen the content consistency of multi-scene videos. Extensive experiments demonstrate that VideoStudio outperforms the SOTA video generation models in terms of visual quality, content consistency, and user preference.
	Source code is available at \url{https://github.com/FuchenUSTC/VideoStudio}.
\end{abstract}

\begin{figure}[tbp]
	\centering
	\includegraphics[width=0.93\textwidth]{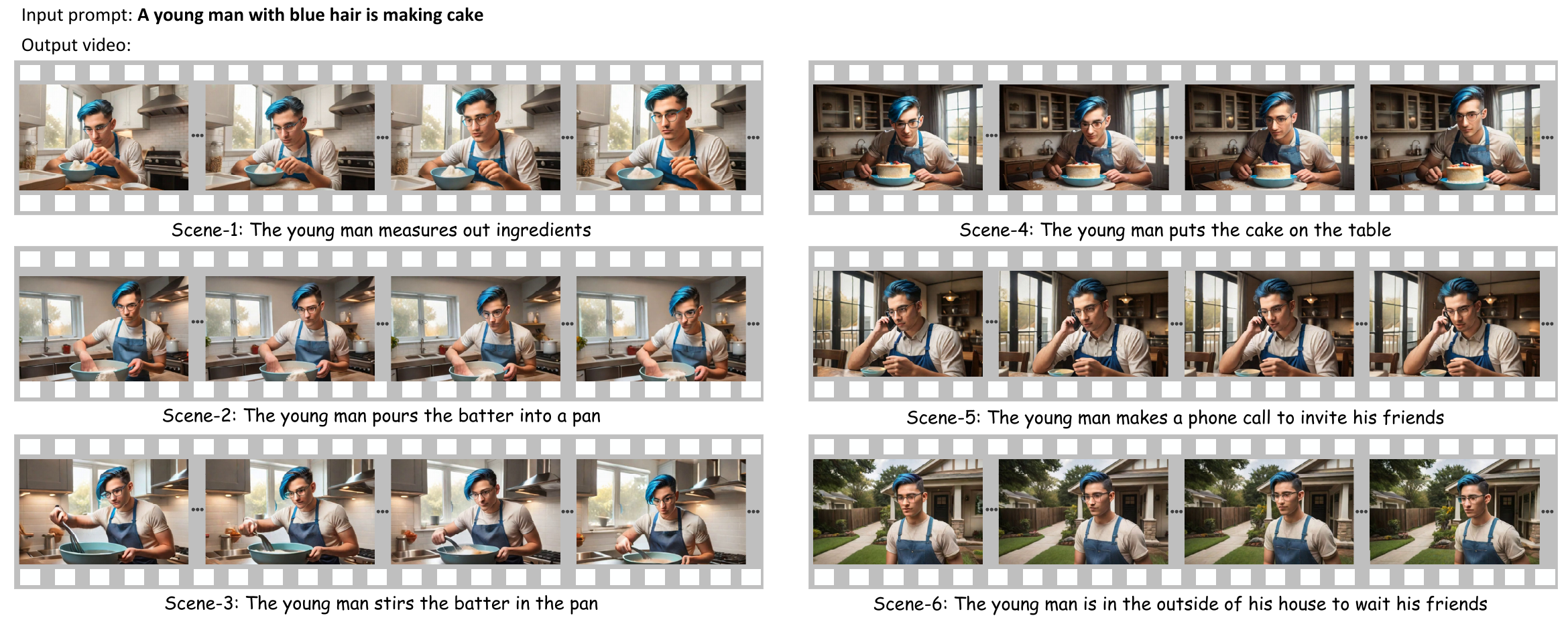}
	\caption{An illustration of prompt and multi-scene video generation by VideoStudio.}
	\label{fig:intro}
\end{figure}

\section{Introduction}
Diffusion Probabilistic Models (DPM) have demonstrated high capability in generating high-quality images~\cite{sohl2015deep, song2019generative, ho2020denoising, nichol2021improved, dhariwal2021diffusion, ho2022classifier, rombach2022high, zhang2023adding, mou2023t2iadapter}. DPM approaches image generation as a multi-step sampling process, involving the use of a denoiser network to progressively transform a Gaussian noise map into an output image. Compared to 2D images, videos have an additional time dimension, which introduces more challenges when extending DPM to video domain. One typical way is to leverage pre-trained text-to-image models to produce video frames~\cite{wu2023tuneavideo, qi2023fatezero, khachatryan2023text2videozero} or utilize a 3D denoiser network learnt on video data to generate a sequence of frames in an end-to-end manner \cite{ho2022imagen, singer2022make, he2022lvdm, guo2023animatediff, luo2023videofusion, blattmann2023align}. Despite having impressive results in the realm of text-to-video generation, most existing works focus on only single-scene videos, featuring one event in a single background. The generation of multi-scene video is still a problem not yet fully explored in the literature.

The difficulty of multi-scene video generation generally originates from two aspects: 1) how to arrange and establish different events in a logical and realistic way for a multi-scene video? 2) how to guarantee the consistency of common entities, e.g., foreground objects or persons, throughout the video? For instance, given an input prompt of ``a young man is making cake,'' a multi-scene video is usually to present the step-by-step procedure of making a cake, including measuring out the ingredients, pouring the ingredients into a pan, cooking the cake, etc. This necessitates a comprehensive understanding and refinement of the prompt. As such, we propose to mitigate the first issue through capitalizing on Large Language Models (LLM) to rewrite the input prompt into multi-scene video script. LLM inherently abstracts quantities of text data on the Web about the input prompt to produce the script, which describes and decomposes the video logically into multiple scenes. To alleviate the second issue, we exploit the common entities to generate reference images as the additional condition to produce each scene video. The reference images, as the link across scenes, effectively align the content consistency within a multi-scene video.

To consolidate the idea, we present a new framework dubbed as \textbf{VideoStudio} for consistent-content and multi-scene video generation. Technically, VideoStudio first transforms the input prompt into a thorough multi-scene video script by using LLM. The script for each scene consists of the descriptive prompt of the event in the scene, a list of foreground objects or persons, the background, and camera movement. VideoStudio then identifies common entities that appear across multiple scenes and requests LLM to enrich each entity. The resultant entity description is fed into a pre-trained Stable Diffusion~\cite{rombach2022high} model to produce a reference image for each entity. Finally, VideoStudio outputs a multi-scene video via involving two diffusion models, i.e., \textbf{VideoStudio-Img} and \textbf{VideoStudio-Vid}. VideoStudio-Img is dedicated to incorporating the descriptive prompt of the event and the reference images of entities in each scene as the condition to generate a scene-reference image. VideoStudio-Vid takes the scene-reference image plus temporal dynamics of the action depicted in the descriptive prompt of the event and camera movement in the script as the inputs and produces a video clip for each scene. 

The main contribution of this work is the proposal of VideoStudio for generating consistent-content and multi-scene videos. The solution also leads to the elegant views of how to use LLM to properly arrange content of multi-scene videos and how to generate visually consistent entities across scenes, which are problems seldom investigated in literature. Extensive experiments conducted on public benchmarks demonstrate that VideoStudio outperforms SOTA video generation models in terms of visual quality, content consistency and user preference.

\section{Related Work}

\textbf{Image generation} is a fundamental challenge of computer vision and has evolved rapidly in the past decade. Recent advances in Diffusion Probabilistic Models (DPM) have led to remarkable improvements in generating high-fidelity images~\cite{blattmann2023align, sohl2015deep, song2019generative, ho2020denoising, nichol2021improved, dhariwal2021diffusion, ho2022classifier, rombach2022high, zhang2023adding, mou2023t2iadapter, nichol2022glide,ramesh2022hierarchical,song2022denoising, lu2022dpmsolver, lu2023dpmsolver}. DPM  is a category of generative models that utilizes a sequential sampling process to convert random Gaussian noise into high-quality images. For example, GLIDE~\cite{nichol2022glide} and DALL-E~2~\cite{ramesh2022hierarchical} exploit the sampling process in the pixel space, conditioned on the text prompt using classifier-free guidance~\cite{ho2022classifier}. Nevertheless, training a powerful denoising network remains challenging due to high computational cost and memory demand associated with sampling at the pixel level. To mitigate this problem, Latent Diffusion Models (LDM)~\cite{rombach2022high} employ sampling in the latent feature space that is established by a pre-trained autoencoder, leading to the improvements on computation efficiency and image quality. Furthermore, the application of DPM is further enhanced by incorporating advanced sampling strategies~\cite{song2022denoising, lu2022dpmsolver, lu2023dpmsolver} and additional control signals~\cite{zhang2023adding,mou2023t2iadapter}.

\textbf{Video generation} is a natural extension of image generation in video domain. The early approaches, e.g., ImagenVideo~\cite{ho2022imagen} and Make-A-Video~\cite{singer2022make}, train video diffusion models in the pixel space, resulting in high computational complexity. Following LDM in image domain, several works ~\cite{luo2023videofusion, blattmann2023align, guo2023animatediff,zhang2024trip,chen2024sateco} propose to exploit the sampling process in the latent feature space for video generation. These works extend the 2D UNet with the transformer layers \cite{li2022CoT,yao2023dual,yao2022wave} to 3D UNet by injecting temporal self-attentions \cite{long2022sifa,long2022dynamic} and/or temporal convolutions \cite{long2019gaussian,long2023BCN}. For instance, Video LDM~\cite{blattmann2023align} and AnimateDiff~\cite{guo2023animatediff} focus on training the injected temporal layers while freezing the spatial layers to preserve the ability of the pre-trained image diffusion model. VideoFusion~\cite{luo2023videofusion} decomposes the 3D noise into a 2D base noise shared across frames and a 3D residual noise, enhancing the correlation between frames. However, the generated videos usually have a limited time duration, typically around 16 frames. Consequently, some recent researches emerge to generate long videos by an extrapolation strategy or hierarchical architecture~\cite{he2022lvdm, liang2022nuwa, villegas2022phenaki, voleti2022mcvd, yin2023nuwa}. In addition, video editing techniques utilize the input video as a condition and generate a video by modifying the style or key object of the input video~\cite{wu2023tuneavideo, shin2023editavideo, qi2023fatezero, hu2023videocontrolnet, geyer2023tokenflow, ouyang2023codef, esser2023structure, he2022lvdm, wang2023videocomposer, yin2023dragnuwa, wang2023genlvideo}.

In short, our work in this paper focuses on consistent-content and multi-scene video generation. The most related work is \cite{lin2023videodirectorgpt}, which aligns the appearance of entities across scenes through the bounding boxes provided by LLM. Ours is different in the way that we explicitly determine the appearance of entities by generating reference images, which serve as a link across scenes and effectively enhance the content consistency within a multi-scene video.

\begin{figure*}[tbp]
	\centering
	\includegraphics[width=0.921\textwidth]{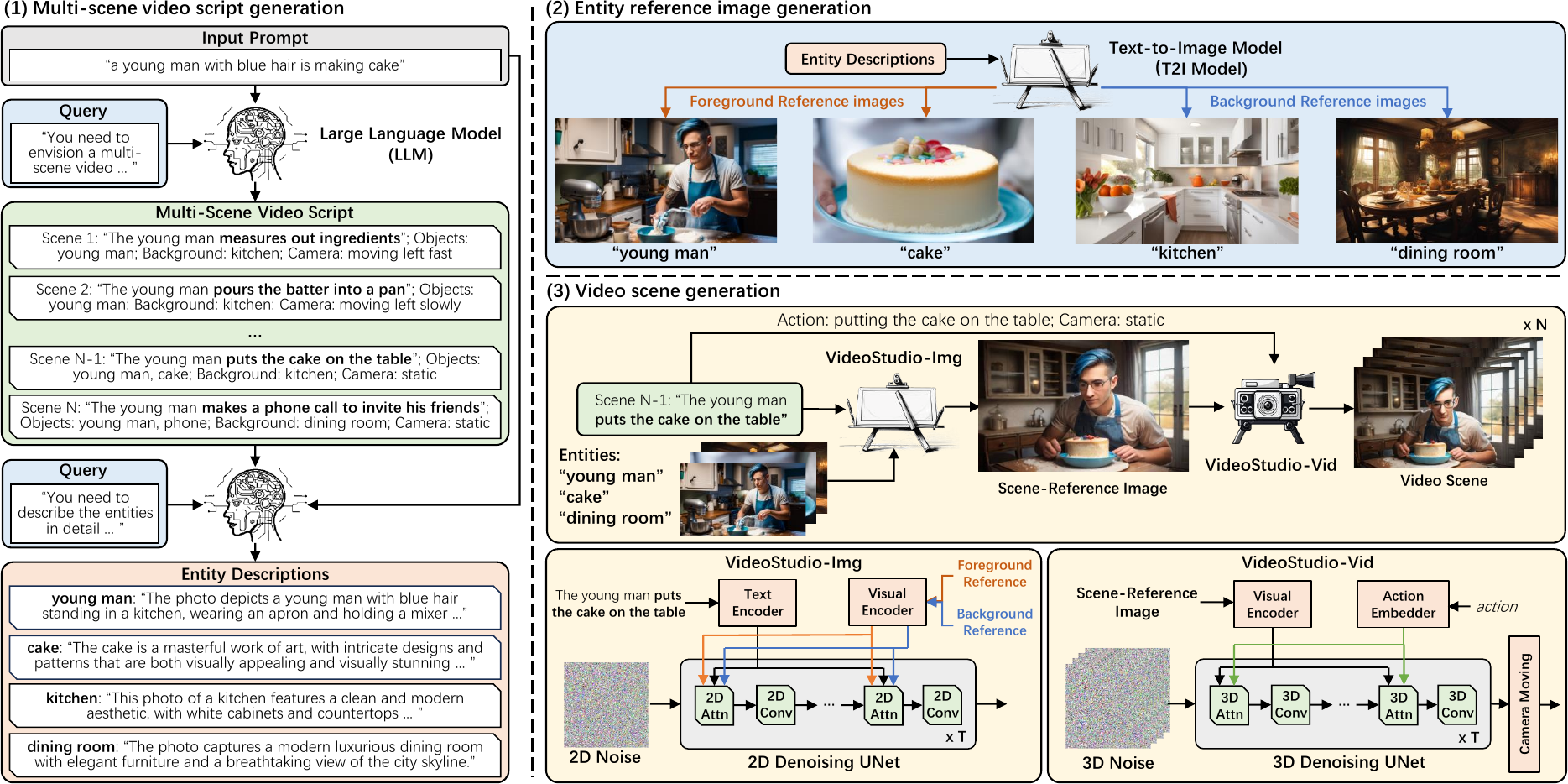}
	\caption{An overview of our VideoStudio framework for consistent-content and multi-scene video generation. VideoStudio consists of three main stages: (1) multi-scene video script generation, (2) entity reference image generation, and (3) video scene generation. In the first stage, LLM is utilized to convert the input prompt into a comprehensive multi-scene script. The script for each scene includes the descriptive prompt of the event in the scene, a list of foreground objects or persons, the background, and camera movement. We then request LLM to detail the common foreground/background entities across scenes. These entity descriptions are fed into a text-to-image (T2I) model to produce reference images in the second stage. Finally, in the third stage, VideoStudio-Img exploits the descriptive prompt of the event and the reference images of entities in each scene as the condition to generate a scene-reference image. VideoStudio-Vid takes the scene-reference image plus temporal dynamics of the action depicted in the descriptive prompt of the event and camera movement in the script as the inputs and produces a video clip for each scene.}
	\label{fig:framework}
\end{figure*}

\section{VideoStudio}
This section presents the proposed VideoStudio framework for consistent-content and multi-scene video generation. Figure~\ref{fig:framework} illustrates an overview of VideoStudio framework, consisting of three main stages: (1) multi-scene video script generation (Sec.~\ref{sec:llm}), (2) entity reference image generation (Sec.~\ref{sec:tem}), and (3) video scene generation (Sec.~\ref{sec:scene}).

\subsection{Multi-Scene Video Script Generation} \label{sec:llm}
As depicted in Figure~\ref{fig:framework}(1), VideoStudio utilizes LLM to convert the input prompt into a comprehensive multi-scene script. In view of its high deployment flexibility and inference efficiency, we use the open-source ChatGLM3-6B model~\cite{zeng2022glm,du2022glm}. The LLM is requested by a pre-defined query, \emph{``You need to envision a multi-scene video and describe each scene ...''}, to treat the input prompt as the theme, logically decompose the video into multiple scenes and generate a script for each scene in the following format:
\begin{equation}
	\small
	\begin{aligned}
		&[\text{Scene $1$: prompt, foreground, background, camera move}]; \\
		&[\text{Scene $2$: prompt, foreground, background, camera move}]; \\
		&~~~~~~~~~~~~~~~~~~~~~~~~~~~~~~~~~~~~~~~~~~~~~... \\
		&[\text{Scene $N$: prompt, foreground, background, camera move}]. \\
	\end{aligned}
\end{equation}
Here $N$ denotes the number of video scenes, which is determined by the LLM. For each scene, the descriptive prompt of the event in the scene, a list of foreground objects or persons, the background, and camera movement are provided. The camera movement is restricted to a close-set of directions \emph{\{static, left, right, up, down, forward, backward\}} and speeds \emph{\{slow, medium, fast\}}.

Next, VideoStudio identifies the common entities, which include foreground objects or persons and background locations. To achieve this, we ask the LLM to assign the common object, person, or background the same name across scenes when generating the video script. Therefore, we strictly match the name of entities and discover the entities that appear in multiple scenes. To further improve the quality of the video script, we employ the capability of the LLM for multi-round dialogue. Specifically, we start the dialogue by asking the LLM to specify the key aspects with respect to the entity, such as \emph{``What are the aspects that should be considered when describing a photo of a young man in detail?''} In the next round of dialogue, we request the LLM to describe the entity from the viewpoints of the given aspects. Moreover, the original prompt is also taken as the input to the LLM to ensure that the essential characteristics, e.g., ``blue hair'' of the young man, are emphasized in entity description generation.  

Please note that the GPT-4 \cite{openai2023gpt4} can also be used for script generation, but it incurs an additional 0.12 USD for the GPT-4 API call per query. In VideoStudio, we leverage the open-source ChatGLM3-6B and perform the inference on our devices to circumvent the need for API call. Nevertheless, the scale of ChatGLM3-6B is much smaller, resulting in unstable outcomes that may deviate from the specified script format. To alleviate this issue, we have empirically abstracted the following principles to enhance the stability of open-source LLM:
\begin{itemize}
	\item[$\bullet$] Before the dialogue starts, we provide comprehensive instructions to the LLM, delineating the additional requirements, specifying the script format, and offering the examples of the expected outputs.
	\item[$\bullet$] For each query, we manually select five in-context examples as the historical context for multi-round dialogue. These examples are very carefully designed to ensure a diverse range of scenes, key objects, and background, and serve to emphasize the required script format for LLM.
	\item[$\bullet$] After each round of dialogue, we verify the output format. If the results are seemingly inappropriate, we re-run the entire script generation stage. Such strategy is simple to implement without requiring any additional expenses.
\end{itemize}
We will provide the full version of our instructions, examples, and queries in the supplementary materials.

\subsection{Entity Reference Image Generation} \label{sec:tem}
In the second stage of VideoStudio, we unify the visual appearance of common entities by explicitly generating a reference image for each entity. The reference images act as the link to cohere the content across scenes. We achieve this by first feeding the entity description into a pre-trained Stable Diffusion model for text-to-image generation. Then, we employ the U$^2$-Net~\cite{Qin_2020_PR} model for salient object detection, and segment the foreground and background areas in each resultant image. By utilizing the segmentation masks, we can further remove the background pixels from the foreground reference image and vice versa, in order to prevent the interference between the foreground and background visual contents in the reference images.

\subsection{Video Scene Generation} \label{sec:scene}
VideoStudio produces a multi-scene video by generating each scene via the diffusion models by taking the reference images, the descriptive prompt of the event and camera movement into account. This stage involves two primary components: the \textbf{VideoStudio-Img}, which utilizes the descriptive prompt of the event and the reference images of entities in each scene as the condition to generate a scene-reference image, and the \textbf{VideoStudio-Vid}, which employs the scene-reference image plus temporal dynamics of the action depicted in the descriptive prompt of the event and camera movement in the script as the inputs and produces a video clip for each scene.

\subsubsection{~}
\begin{figure}[tbp]
	\begin{minipage}[t]{0.5\linewidth}
		\centering
		\includegraphics[width=0.77\textwidth]{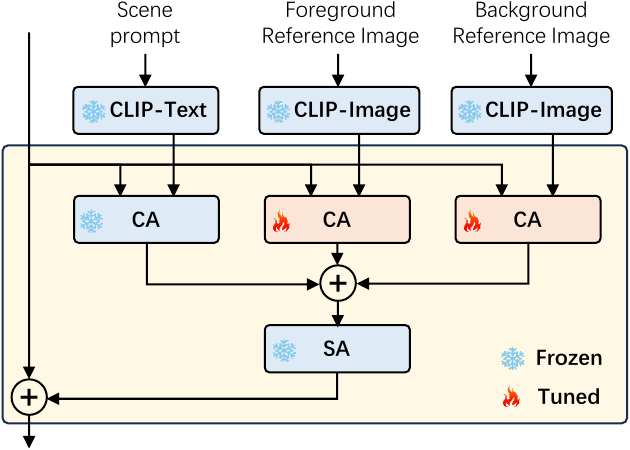}
		\subcaption{}
		\label{fig:attn1}
	\end{minipage}
	\begin{minipage}[t]{0.5\linewidth}
		\centering
		\includegraphics[width=0.77\textwidth]{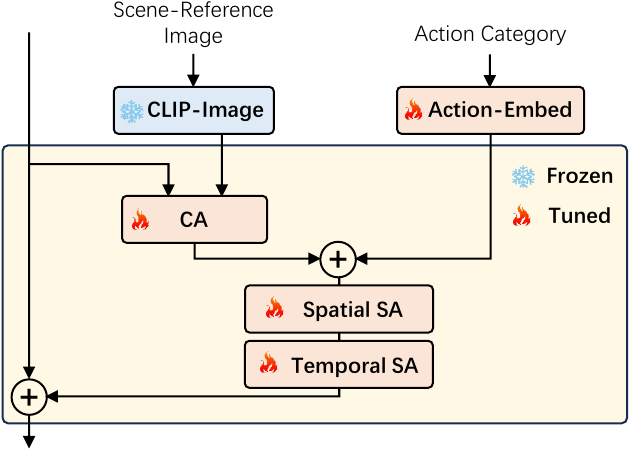}
		\subcaption{}
		\label{fig:attn2}
	\end{minipage}
	\caption{Diagram illustrations of (a) attention module in the VideoStudio-Img which takes the scene prompt and foreground/background reference images as the inputs and (b) attention module in the VideoStudio-Vid conditioning on the scene-reference image and the described action category.}
\end{figure}

The \textbf{VideoStudio-Img} component aims to generate a scene-reference image conditioning on the event prompt and entity reference images for each scene. To accomplish this, we remold the Stable Diffusion architecture by replacing the original attention module with a novel attention module that can handle three contexts: the text prompt, foreground reference image, and background reference image. As depicted in Figure~\ref{fig:attn1}, we utilize text and visual encoder of a pre-trained CLIP model to extract the sequential text feature $y_t \in \mathbb{R}^{L_t \times C_t}$ and local image features $y_f \in \mathbb{R}^{L_f \times C_f}$ and $y_b \in \mathbb{R}^{L_b \times C_b}$ for the prompt, foreground reference image, and background reference image, respectively. Here, $L$ and $C$ denote the length and the channels of the feature sequence. For the case of multiple foregrounds in one scene, we concatenate the features from all foreground reference images along the length dimension. Given the input feature $\bm{x}$, the outputs $\bm{z}$ of the attention are computed as
\begin{equation}
	\small
	\begin{aligned}
		\bm{y} &= \text{CA}_1(\bm{x}, y_t)+\text{CA}_2(\bm{x}, y_f)+\text{CA}_3(\bm{x}, y_b), \\
		\bm{z} &= \bm{x} + \text{SA}(\bm{y}),
	\end{aligned}
\end{equation}
where CA$_1$ and SA are the cross-attention and self-attention modules, respectively, in the original Stable Diffusion architecture. We add two additional cross-attention modules, CA$_2$ and CA$_3$, which leverage the guidance provided by entity reference images. Moreover, we propose to optimize the parameters of CA$_2$ and CA$_3$ while freezing the other parts of the network.

\subsubsection{~}
The \textbf{VideoStudio-Vid} is a video diffusion model that employs the scene-reference image, the action described in the prompt of the event, and camera movement in the script as the inputs. Particularly, we start by extending the Stable Diffusion model to a spatio-temporal form and replacing the original attention module with a new one that is conditioned on the scene-reference image and action category, as shown in Figure~\ref{fig:attn2}. Taking 400 action categories in Kinetics~\cite{carreira2017quo} as an action vocabulary, an indicator vector $y_a\in [0,1]^{400}$ is built to infer if each action in the vocabulary exists in the scene prompt and subsequently converted into feature space using a linear embedding $f$. For the scene-reference image, we use the visual encoder of CLIP to extract the image feature $y_s \in \mathbb{R}^{L_s \times C_s}$, which is then fed into the cross-attention operation. The original self-attention is decomposed into a spatial self-attention (Spatial SA) and a temporal self-attention (Temporal SA), which operate self-attention solely on spatial and temporal dimension, respectively, to reduce computations. Hence, given the input feature $\bm{x}$, the attention module is formulated as
\begin{equation}
	\small
	\begin{aligned}
		\bm{y} &= \text{CA}(\bm{x}, y_s)+ f(y_a), \\
		\bm{z} &= \bm{x} + \text{Temporal SA}(\text{Spatial SA}(\bm{y})).
	\end{aligned}
\end{equation}
Moreover, we further inject several temporal convolutions behind each spatial convolution into the Stable Diffusion architecture, to better capture temporal dependencies in image-to-video generation.

To reflect the camera movement stated by the script in the generated video, we uniquely modify the frames in the intermediate step of sampling process by warping the neighboring frames based on the camera moving direction and speed. We execute this adjustment after the first $T_m$ DDIM sampling steps, followed by continuing the sampling process. Such modification ensures that the resultant video clip maintains the same camera movement as we warp the intermediate frames. In general, setting a small $T_m$ for early modification may not effectively control the camera movement, while a late modification may affect the visual quality of the output videos. In practice, we observe that $T_m$=5 provides a good trade-off. We will detail the formulation of the modification process and the ablation study of the step $T_m$ in our supplementary materials.

\section{Experiments}
\subsection{Datasets}
Our VideoStudio framework is trained on three large-scale datasets: LAION-2B~\cite{schuhmann2022laion}, WebVid-10M~\cite{bain2021frozen} and HD-VG-130M~\cite{wang2023videofactory}. The LAION-5B is one of the largest text-image dataset consisting of around 5 billion text-image pairs. To train VideoStudio-Img, We utilize a subset, namely \textbf{LAION-2B}, which focuses on the text prompts in English. The \textbf{WebVid-10M} and \textbf{HD-VG-130M} are the large-scale single-scene video datasets, containing approximately 10M and 130M text-video pairs, respectively.
VideoStudio-Vid is trained on the combination of WebVid-10M and a randomly chosen 20M subset from HD-VG-130M.

To evaluate video generation, we select the text prompts from three video datasets, i.e., MSR-VTT~\cite{xu2016msr}, ActivityNet Captions~\cite{krishna2017dense} and Coref-SV~\cite{lin2023videodirectorgpt}. The first one provides the single-scene prompts, while the remaining two datasets comprise multi-scene prompts. The \textbf{MSR-VTT} consists of 10K web video clips, each annotated with approximate 20 natural sentences. We utilize the text annotation of validation videos to serve as single-scene prompts in our evaluation. The \textbf{ActivityNet Captions} dataset is a multi-event video dataset designed for dense-captioning tasks. Following \cite{lin2023videodirectorgpt}, we randomly sample 165 videos from the validation set and exploit the event captions as the multi-scene prompts. The \textbf{Coref-SV} is a multi-scene description dataset, which was constructed by replacing the subject of multi-scene paragraphs in Pororo-SV dataset~\cite{kim2017deepstory,li2019storygan}. Coref-SV samples 10 episodes from the Pororo-SV dataset and replaces the subject with 10 real-world entities, resulting in 100 multi-scene prompts.

\subsection{Evaluation Metrics}
For the video generation task, we adopt five evaluation metrics. To assess the visual quality of the generated videos, we utilize the average of the per-frame Fr\'{e}chet Inception Distance (\textbf{FID})~\cite{heusel2017gans} and the clip-level Fr\'{e}chet Video Distance (\textbf{FVD})~\cite{unterthiner2019fvd}, both of which are commonly used metrics. We also employ the \textbf{CLIPSIM}~\cite{wu2021godiva} metric to evaluate the alignment between the generated frames and the input prompt. To verify the content consistency, we calculate frame consistency (\textbf{Frame Consis.}) by determining the CLIP-similarity between consecutive frames, serving as an intra-scene consistency measure. Additionally, we employ the Grounding-DINO detector~\cite{liu2023grounding} to detect common objects across scenes and then calculate the CLIP-similarity between the common objects appeared in different scenes, achieving cross-scene consistency (\textbf{Scene Consis.}).

\subsection{Implementation Details}
We implement the proposed VideoStudio using the Diffusers codebase on the platform of PyTorch.

\textbf{Training stage of VideoStudio-Img.} VideoStudio-Img is originated from the Stable Diffusion v2.1 model by incorporating two additional cross-attention modules. These modules are initialized from scratch and trained on the text-image pairs from LAION-2B dataset, while other parts of the network are frozen. For each image, we randomly sample a 512$\times$512 patch cropped from the original image, and utilize the U$^2$-Net model to segment the foreground area of each patch. The isolated foreground and background areas serve as the foreground and background reference images, respectively, to guide the generation of the input patch. We set each minibatch as 512 patches that are processed on 64 A100 GPUs in parallel. The parameters of the model are optimized by AdamW optimizer with a fixed learning rate of $1\times 10^{-4}$ for 20K iterations.

\textbf{Training stage of VideoStudio-Vid.} VideoStudio-Vid is developed based on the Stable Diffusion XL architecture by inserting temporal attentions and temporal convolutions. The training is carried out on the WebVid-10M and HD-VG-130M datasets. For each video, we randomly sample a 16-frame clip with the resolution of 320$\times$512 and an FPS of 8. The middle frame of the clip is utilized as the scene-reference image. Each minibatch consists of 128 video clips implemented on 64 A100 GPUs in parallel. We utilize the AdamW optimizer with a fixed learning rate of $3\times 10^{-6}$ for 480K iterations.

\subsection{Experimental Analysis of VideoStudio}
\begin{figure*}
	\begin{minipage}{0.49\linewidth}
		\centering
		\captionsetup{type=table}
		\caption{\small Performance comparisons of IP-Adapter \cite{ye2023ipadapter} and VideoStudio-Img variants with different input references on the MSR-VTT validation set.}
		\setlength{\tabcolsep}{0.2cm}\resizebox{0.99\textwidth}{!}{
			\begin{tabular}{c c|c|c|c}
				\toprule
				\multicolumn{2}{c|}{{\textbf{Input References}}} & \multirow{2}{*}{{\textbf{FG-SIM}}} & \multirow{2}{*}{{\textbf{BG-SIM}}} & \multirow{2}{*}{{\textbf{CLIPSIM}}} \\
				FG Ref. & BG Ref. &  &  \\ \midrule
				\multicolumn{2}{c|}{w/o Ref.}		  	  &  0.5162     &  0.4131  & {0.3001} \\ \midrule
				\multicolumn{5}{c}{IP-Adapter \cite{ye2023ipadapter}} \\ \midrule
				\checkmark   &     &  0.7116    &  0.4035  & 0.2910   \\
				& \checkmark       &  0.5128    &  0.5059  & 0.2954   \\ \midrule
				\multicolumn{5}{c}{VideoStudio-Img} \\ \midrule
				\checkmark   &                 			  &  \underline{0.7919}    &  0.4393  & 0.2982   \\
				& \checkmark     					      &  0.5362     &  \underline{0.5742}  & \underline{0.3002}   \\ \midrule
				\checkmark   & \checkmark      		      &  \textbf{0.8102}   &  \textbf{0.5861} & \textbf{0.3023}   \\ \bottomrule
			\end{tabular}
		}
		\label{tab:ab.img}
	\end{minipage}
	\begin{minipage}{0.49\linewidth}
		\centering
		\caption{Examples of the foreground and background reference images and the generated scene-reference image by the VideoStudio-Img variants.}
		\includegraphics[width=1.0\textwidth]{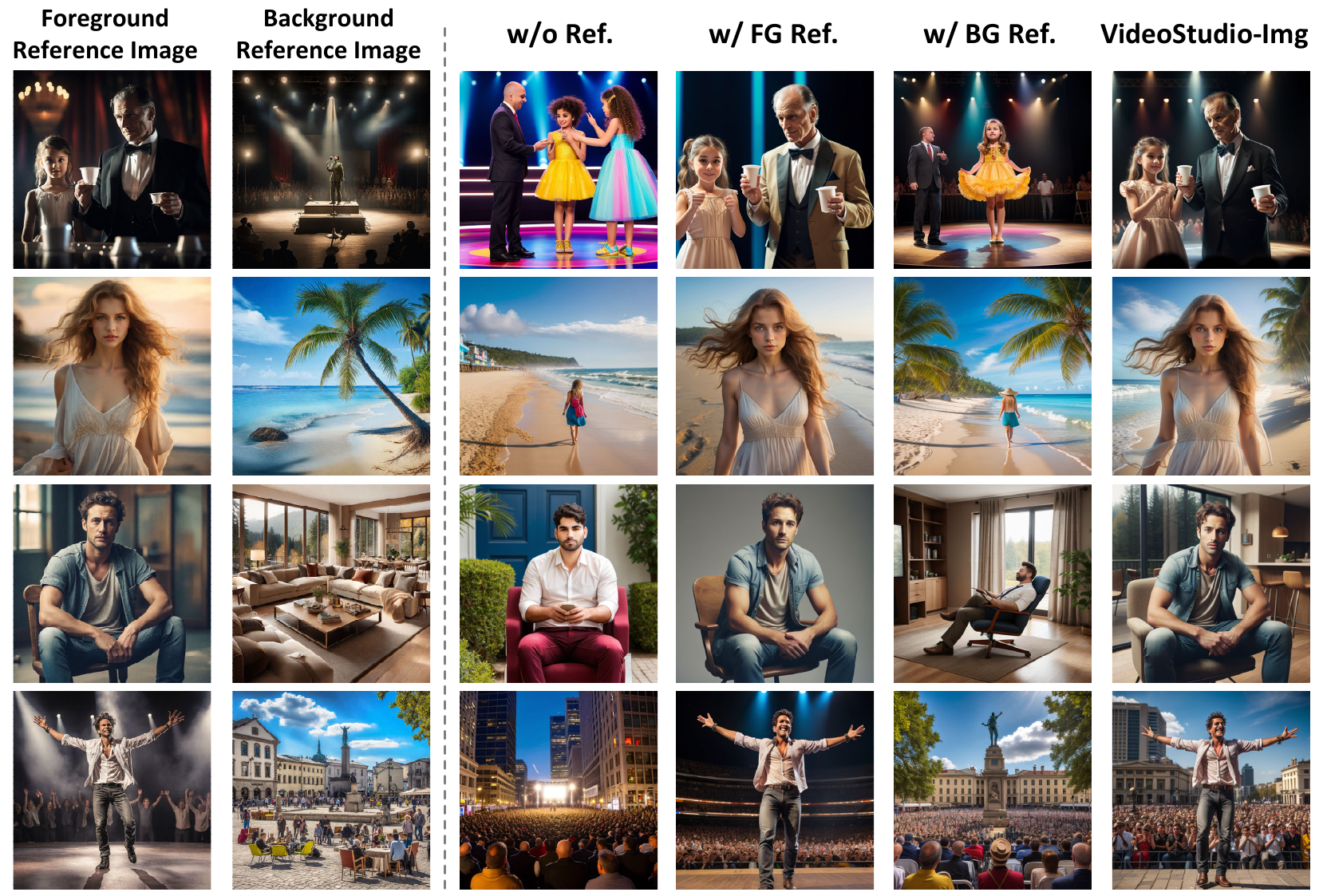}
		\label{fig:img}
	\end{minipage}
\end{figure*}

\textbf{Evaluation on VideoStudio-Img.} 
We first verify the efficacy of VideoStudio-Img in aligning with the input entity reference images. To this end, we take the prompts from MSR-VTT validation set. The input foreground and background reference images are produced by using LLM and Stable Diffusion model. We validate the generated images on the measure of foreground similarity (\textbf{FG-SIM}) and background similarity (\textbf{BG-SIM}), which are the CLIP-similarity values with the foreground and background reference images, respectively. 
Table~\ref{tab:ab.img} lists the performance comparisons of IP-Adapter \cite{ye2023ipadapter} and different VideoStudio-Img variants by leveraging different input references. 
\begin{table}[t]
	\centering
	\caption{Performance comparisons for single-scene video generation with real frame as scene-reference image on WebVid-10M.}
	\setlength{\tabcolsep}{1.7cm}\resizebox{0.9\linewidth}{!}{
		\begin{tabular}{l |c|c}
			\toprule
			\textbf{Approach}       & \textbf{FVD} ($\downarrow$) & \textbf{Frame Consis.} ($\uparrow$)  \\ \midrule
			RF+VideoCrafter \cite{chen2023videocrafter1} 	& 293.3  & 97.9     \\
			RF+I2VGen-XL \cite{zhang2023i2vgen}			 	& 254.9  & 97.6      \\
			RF+VideoComposer \cite{wang2023videocomposer}   & 231.0  & 95.9  \\ 
			RF+DynamiCrafter \cite{xing2023dynamicrafter}   & 176.8  & 97.5      \\ 
			RF+SVD \cite{blattmann2023svd}					& 153.0  & 98.7  \\ \midrule
			RF+VideoStudio-Vid$^{-}$ 				   	    & 157.3  & 98.5  \\
			RF+VideoStudio-Vid 					        & \textbf{116.5} & \textbf{98.8}  \\
			\bottomrule
		\end{tabular}
	}
	\label{tab:vid.webvid}
\end{table}
Specifically, the use of foreground/background reference image as guidance leads to higher FG-SIM/BG-SIM values comparing to IP-Adapter or not leveraging reference images. 
Though both of IP-Adapter and VideoStudio-Img exploit additional cross-attention to maintain visual contents in image diffusion, our VideoStudio-Img is devised for a more complex scenario to specify foreground objects and background.
There are two major differences: 1) We pre-segment the foreground/background of the reference images to avoid the visual content interference; 2) IP-Adapter extracts global image features from CLIP, while ours utilizes local image tokens from CLIP to improve spatial discrimination in local regions.
As indicated by the results, emphasizing the feature learning of local region on the more clean (masked) foreground/background reference image does benefit the visual alignment.
Furthermore, the combination of both reference images achieves the highest FG-SIM of $0.8102$ and BG-SIM of $0.5861$. 
Figure~\ref{fig:img} showcases four generated images by different VideoStudio-Img variants with various reference images. The results demonstrate the advantage of VideoStudio-Img to align with the visual contents in the entity reference images.

\textbf{Evaluation on VideoStudio-Vid.} Next, we assess the visual quality of the single-scene videos generated by VideoStudio-Vid. We exploit the real frame from the WebVid-10M validation set as the scene-reference image irrespective of the generation quality, and produce a video using the corresponding text prompt, which is referred to as RF+VideoStudio-Vid. 
We compare our proposal with five image-to-video diffusion models and one variant of VideoStudio-Vid, i.e., RF+VideoStudio-Vid$^-$, which disables the action guidance in VideoStudio-Vid.
Table~\ref{tab:vid.webvid} presents the performance comparisons for single-scene video generation on the WebVid-10M dataset. 
With the same scene-reference images, VideoStudio-Vid$^{-}$ outperforms most image-to-video approaches and obtains comparable FVD performance with the strong baseline SVD. 
The competitive result is attributed to the deep network architecture and large-scale training set.
The performance is further enhanced to $116.5$ FVD and $98.8$ frame consistency by RF+VideoStudio-Vid, verifying the superiority of involving action category guidance to improve visual quality and intra-scene consistency.

\begin{table}[t]
	\centering
	\caption{Performance comparisons for single-scene video generation on MSR-VTT validation set. RF indicates whether to utilize the real frame as the reference.}
	\setlength{\tabcolsep}{1.7cm}\resizebox{0.95\linewidth}{!}{
		\begin{tabular}{l |c|c|c}
			\toprule
			\textbf{Approach} & \textbf{RF}  & \textbf{FID} ($\downarrow$)  & \textbf{FVD} ($\downarrow$)    \\
			\midrule
			CogVideo \cite{2022cogvideo} 		   &  & 23.6 & - \\
			MagicVideo \cite{2023magicvideo}	   &  &  -	& 998 \\
			Make-A-Video \cite{singer2022make}     &  & 13.2 & -    \\ 
			VideoComposer \cite{wang2023videocomposer} &  & - 	& 580 \\
			VideoDirectorGPT \cite{lin2023videodirectorgpt}     &  & 12.2 & {550} \\
			ModelScopeT2V \cite{2023modelscope}    &  & \textbf{11.1} & {550} \\
			SD+VideoStudio-Vid 				   &  & {11.9} & \textbf{381} \\ 
			\midrule
			RF+VideoCrafter \cite{chen2023videocrafter1}  & \checkmark & 45.0 & 339 \\
			RF+I2VGen-XL \cite{zhang2023i2vgen}			  & \checkmark & 37.4 & 264 \\
			RF+VideoComposer \cite{wang2023videocomposer} & \checkmark & 31.3 & 208 \\ 
			RF+DynamiCrafter \cite{xing2023dynamicrafter} & \checkmark & 26.1 & 196 \\
			RF+SVD \cite{blattmann2023svd}				  & \checkmark & 15.3 & 172  \\  
			RF+VideoStudio-Vid 					      & \checkmark & \textbf{10.8} & \textbf{133} \\
			\bottomrule
		\end{tabular}
	}
	\label{tab:vid.msrvtt}
\end{table}

\begin{figure*}[tbp]
	\centering
	\includegraphics[width=0.95\textwidth]{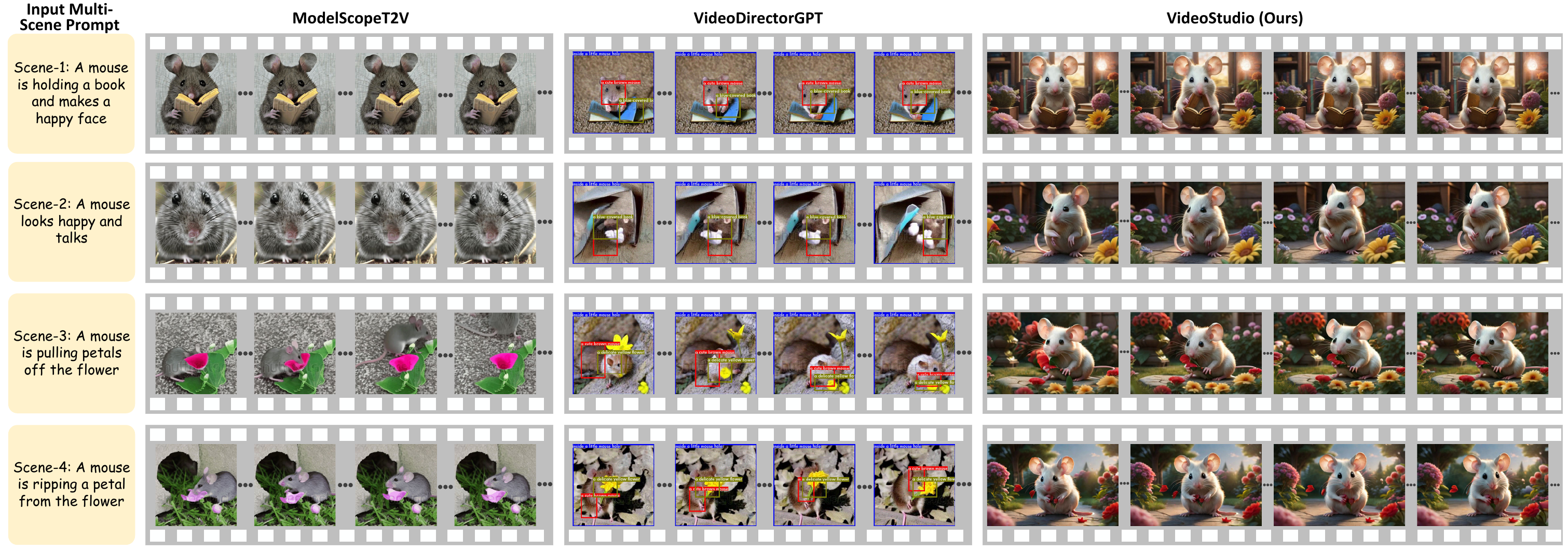}
	\caption{Examples of generated multi-scene videos by ModelScopeT2V~\cite{2023modelscope}, VideoDirectorGPT~\cite{lin2023videodirectorgpt} and our VideoStudio utilizing a multi-scene prompt from the Coref-SV dataset. For each video, only the first four scenes are given. The results of VideoDirectorGPT are provided in the project webpage and thus with bounding box annotation.}
	\label{fig:sota}
\end{figure*}

Similar performance trends are observed on MSR-VTT dataset, as summarized in Table~\ref{tab:vid.msrvtt}. The methods in this table are grouped into two categories: the methods with or without real frame (RF) as reference. To compare with the generation models without RF, we develop a two-step solution that first generates the scene-reference image by Stable Diffusion, and then converts the image into a video clip by VideoStudio-Vid, which is denoted as \textbf{SD+VideoStudio-Vid}. Specifically, VideoStudio-Vid attains the best FVD on both settings with and without a real frame as reference. SD+VideoStudio-Vid is slightly inferior to ModelScopeT2V in FID. We speculate that this may be the result of not optimizing Stable Diffusion on video frames, resulting in poorer frame quality against ModelScopeT2V. Nevertheless, SD+VideoStudio-Vid apparently surpasses ModelScopeT2V in FVD, validating the video-level quality by VideoStudio-Vid.

To evaluate the effectiveness of the action category condition for motion generation, we additionally implement an ablation study on the recent released VBench \cite{huang2023vbench} benchmark.
We measure the \textbf{action score} in VBench to assess whether human subjects can accurately execute the specific action mentioned in the text prompts.
By using the action category as the condition in video diffusion, the action score of VideoStudio-Vid is improved from $90.3\%$ to $96.5\%$, indicating the efficacy of action category condition to emphasize motion patterns.

\begin{table}[t]
	\centering
	\caption{Performance comparisons for multi-scene video generation on ActivityNet Captions dataset.}
	\setlength{\tabcolsep}{1.4cm}\resizebox{0.91\linewidth}{!}{
		\begin{tabular}{l |c|c|c}
			\toprule
			\textbf{Approach} & \textbf{FID} ($\downarrow$)  & \textbf{FVD} ($\downarrow$) & \textbf{Scene Consis.} ($\uparrow$)     \\
			\midrule
			ModelScopeT2V \cite{2023modelscope}       			    & 18.1 & 980 & 46.0 \\
			VideoDirectorGPT \cite{lin2023videodirectorgpt}         & 16.5 & 805 & 64.8 \\ \midrule
			VideoStudio w/o Ref.                            		& 17.3 & 624 & 50.8 \\
			VideoStudio 					           				& \textbf{13.2} & \textbf{395} & \textbf{75.1}  \\
			\bottomrule
		\end{tabular}
	}
	\label{tab:scene.anet}
\end{table}

\begin{figure}[tbp]
	\centering
	\includegraphics[width=0.91\textwidth]{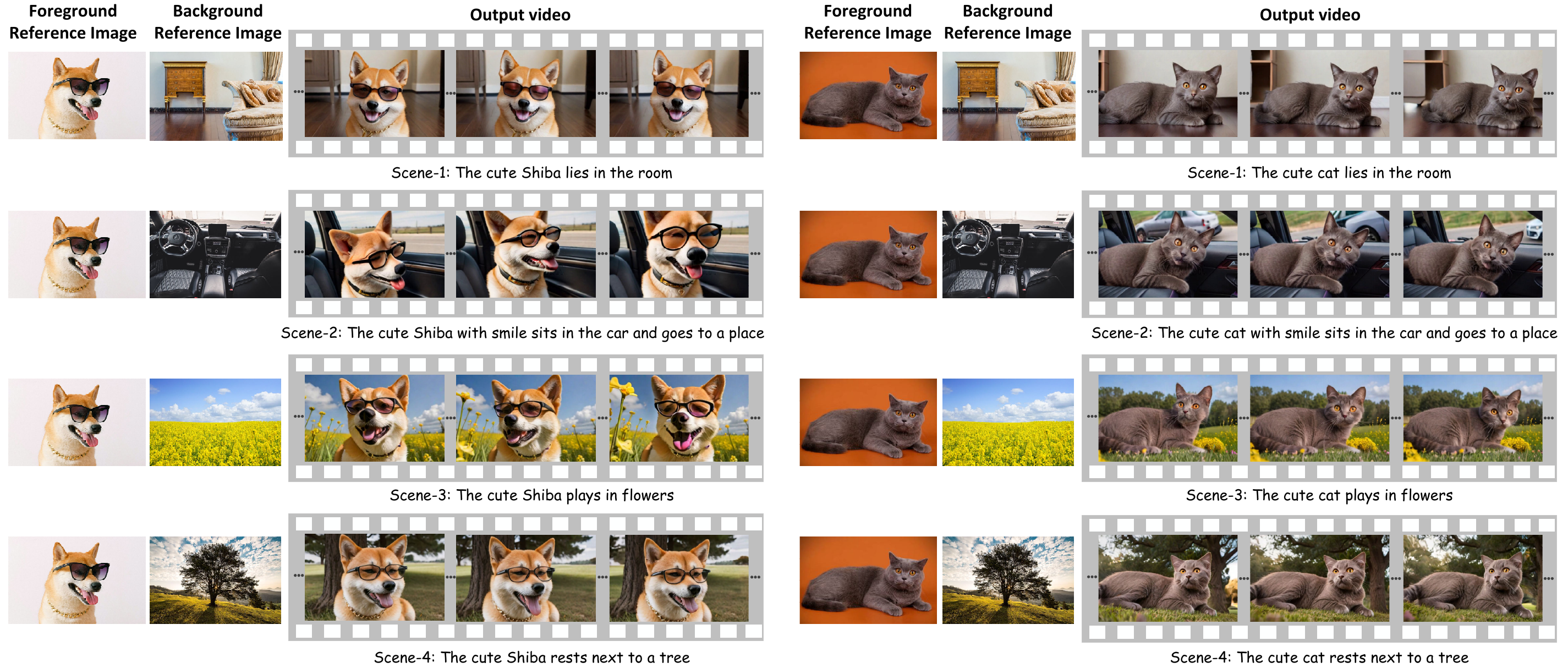}
	\caption{Two examples of generated multi-scene videos by our VideoStudio using the real images as entity reference images.}
	\label{fig:real}
\end{figure}

\begin{table}[t]
	\centering
	\caption{Performance comparisons for multi-scene video generation on Coref-SV.}
	\setlength{\tabcolsep}{1.4cm}\resizebox{0.91\linewidth}{!}{
		\begin{tabular}{l|c|c}
			\toprule
			\textbf{Approach} & \textbf{CLIPSIM} ($\uparrow$) & \textbf{Scene Consis.} ($\uparrow$)     \\
			\midrule
			ModelScopeT2V \cite{2023modelscope}    					& 0.3021 & 37.9 \\
			VideoDirectorGPT \cite{lin2023videodirectorgpt}         & - & 42.8 \\ \midrule
			VideoStudio w/o Ref.                            		& 0.3103 & 40.9 \\
			VideoStudio 					           				& \textbf{0.3304} & \textbf{77.3}  \\
			\bottomrule
		\end{tabular}
	}
	\label{tab:scene.coref}
\end{table}

\subsection{Evaluations on Multi-Scene Video Generation}
We validate VideoStudio for multi-scene video generation on ActivityNet Captions and Coref-SV datasets. Both of the datasets consist of multi-scene prompts, which necessitate the LLM to write the video script based on the given prompt of each scene. We compare with three approaches: ModelScopeT2V, VideoDirectorGPT and VideoStudio w/o Ref. by disabling the reference images in VideoStudio. Table~\ref{tab:scene.anet} details the performance comparisons on ActivityNet Captions. As indicated by the results in the table, VideoStudio exhibits superior visual quality and better cross-scene consistency. Specifically, VideoStudio surpasses VideoStudio w/o Ref. by $24.3$ scene consistency, which essentially verifies the effectiveness of incorporating entity reference images. Moreover, VideoStudio leads to $10.3$ and $29.1$ improvements in scene consistency over VideoDirectorGPT and ModelScopeT2V, respectively. Similar results are also observed on Coref-SV dataset, as summarized in Table~\ref{tab:scene.coref}. Note that as Coref-SV only offers prompts without the corresponding videos, FID and FVD cannot be measured for this case. As shown in the table, VideoStudio again achieves the highest cross-scene consistency of $77.3$, making an absolute improvement of $39.4$ and $34.5$ over ModelScopeT2V and VideoDirectorGPT. Figure~\ref{fig:sota} showcases an example of generated four-scene videos by different approaches on Coref-SV, manifesting the ability of VideoStudio on generating visually similar entities (e.g., mouse/garden) across scenes. Figure~\ref{fig:real} further shows two examples of multi-scene video generation by VideoStudio \textbf{using the real images as entity reference images}, which demonstrates the potential of VideoStudio in customizing the generated objects or environments.

\begin{figure*}[tbp]
	\centering
	\includegraphics[width=0.95\textwidth]{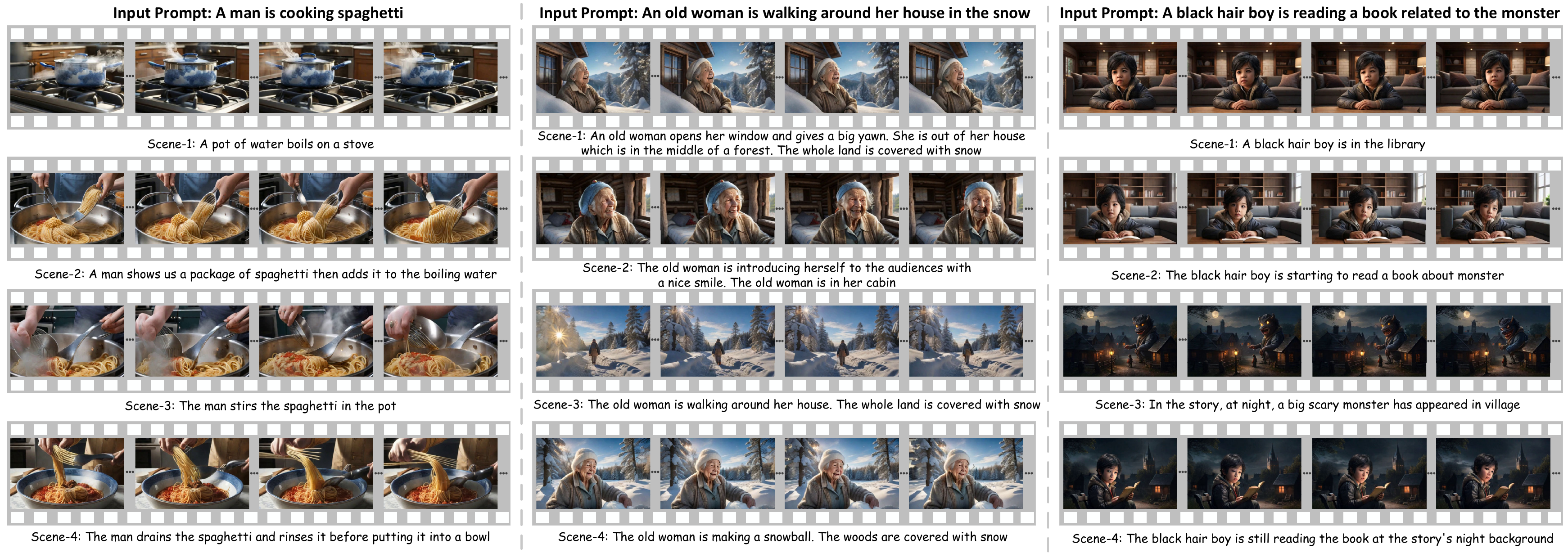}
	\caption{Examples of generated multi-scene videos by VideoStudio on MSR-VTT. For each video, only the first four scenes are given.}
	\label{fig:pipe}
\end{figure*}

\begin{table}[t]
	\centering
	\caption{The user study on three criteria: visual quality (VQ), logical coherence (LC) and content consistency (CC).}
	\setlength{\tabcolsep}{1.4cm}\resizebox{0.85\linewidth}{!}{
		\begin{tabular}{l|ccc}
			\toprule
			\textbf{Approach} & \textbf{VQ} ($\downarrow$) & \textbf{LC} ($\downarrow$) & \textbf{CC} ($\downarrow$)    \\
			\midrule
			ModelScopeT2V w/o LLM & 4.5 & 4.7 & 3.9 \\
			ModelScopeT2V w/ LLM & 4.5 & 3.8 & 4.2 \\ \midrule
			VideoStudio  w/o Ref. w/o LLM & 2.0 & 3.0 & 2.3 \\ 
			VideoStudio w/o Ref. & 2.4 & 2.3 & 3.4 \\
			VideoStudio & \textbf{1.6} & \textbf{1.2} & \textbf{1.2} \\
			\bottomrule
		\end{tabular}
	}
	\label{tab:he}
\end{table}

\subsection{Human Evaluation}
\textbf{Multi-Scene Video Quality.} 
In this section, we conduct a human study to evaluate the entire process of generating multi-scene video from a single prompt. We compare our VideoStudio with four approaches: \textbf{ModelScopeT2V w/o LLM} and \textbf{VideoStudio w/o Ref. w/o LLM} to generate five scenes by duplicating the input prompt, \textbf{ModelScopeT2V w/ LLM} and \textbf{VideoStudio w/o Ref.} to utilize LLM to provide video script as described in Sec. \ref{sec:llm} while generate each scene individually. We invite 12 evaluators and randomly select $100$ prompts from MSR-VTT validation set for human evaluation. We show all the evaluators the five videos generated by each approach plus the given prompt and ask them to rank the five videos from 1 to 5 (good to bad) with respect to the three criteria: visual quality (\textbf{VQ}), logical coherence (\textbf{LC}) and content consistency (\textbf{CC}). For each approach, we average the ranking on each criterion of all the generated videos. As indicated by the results in Table~\ref{tab:he}, the study proves the impact of LLM in generating video script and entity reference images to improve logical coherence and content consistency, respectively. Figure~\ref{fig:pipe} illustrates the examples of the generated multi-scene videos by our VideoStudio.

\begin{table}[t]
	\centering
	\caption{\small User preferences on script/videos by using different LLMs in VideoStudio.}
	\setlength{\tabcolsep}{1.3cm}\resizebox{0.85\linewidth}{!}{
		\begin{tabular}{@{~~~~~~~~~~}l@{~~~~~~~~~~}|@{~~~~~~~~~~}c@{~~~~~~~~~~}c@{~~~~~~~~~~}c@{~~~~~~~~~~}}
			\toprule
			& ChatGLM3-6B \cite{du2022glm} & GPT-4 \cite{openai2023gpt4} & Tie  \\ \midrule
			Video Script & 25\% & 37\% & 38\% \\
			Multi-Scene Video & 20\% & 21\% & 59\% \\
			\bottomrule
	\end{tabular}}
	\label{tab:llm}
\end{table}

\textbf{Different LLMs.}
To further investigate the effectiveness of different LLMs for multi-scene video generation, we carried out an ablation study on a variant of VideoStudio with GPT-4 \cite{openai2023gpt4} in Table \ref{tab:llm}. 
Evaluators vote on the preferring video text script by using ChatGLM3-6B and GPT-4, and the corresponding multi-scene videos generated by VideoStudio. ``Tie'' refers to a close preference. The results indicate that the video script generated by GPT-4 is of higher quality than ChatGLM3-6B. This is not surprising given the significantly larger parameters of GPT-4 ($\sim$1T v.s. 6B). Nevertheless, the voting on multi-scene videos is comparable, showing that the use of an open-source LLM does not affect video quality much.
Our exploitation of open-source LLM leads to an elegant view of how responses of light-weight LLM could be improved for video script generation.

\section{Conclusions}
We have presented a new VideoStudio framework for consistent-content and multi-scene video generation. VideoStudio involves LLM to benefit from the logical knowledge learnt behind and rewrite the input prompt into a multi-scene video script. Then, VideoStudio identifies common entities throughout the script and generates a reference image for each entity, which serves as the link across scenes to ensure the appearance consistency. To produce a multi-scene video, VideoStudio devises two diffusion models of VideoStudio-Img and VideoStudio-Vid. VideoStudio-Img creates a scene-reference image for each scene based on the corresponding event prompt and entity reference images. VideoStudio-Vid converts the scene-reference image into a video clip conditioning on the specific action and camera movement.
Extensive evaluations on four video benchmarks demonstrate the superior visual quality and content consistency by VideoStudio over SOTA models.


%
%
\bibliographystyle{splncs04}
\bibliography{egbib}

\input{sup/VideoStudio_sup}

\end{document}

%% file: sup/VideoStudio_sup.tex
\title{VideoStudio: Generating Consistent-Content and Multi-Scene Videos \\
	   --- ECCV 2024 Supplementary Material}

\titlerunning{VideoStudio}

\author{Fuchen Long \and
	Zhaofan Qiu \and
	Ting Yao \and
	Tao Mei}


\authorrunning{Fuchen Long, Zhaofan Qiu, Ting Yao and Tao Mei}

\institute{HiDream.ai Inc. \\
	\email{\{longfuchen, qiuzhaofan, tiyao, tmei\}@hidream.ai}
}

\maketitle

The supplementary material contains: 1) the instructions of LLM; 2) the implementation details of VideoStudio-Img; 3) the implementation details of VideoStudio-Vid; 4) performance contribution of VideoStudio; 5) more video examples generated by VideoStudio; 6) a video demo for VideoStudio.

\section{Instructions of LLM}
\begin{figure*}[tbp]
	\centering
	\includegraphics[width=0.96\textwidth]{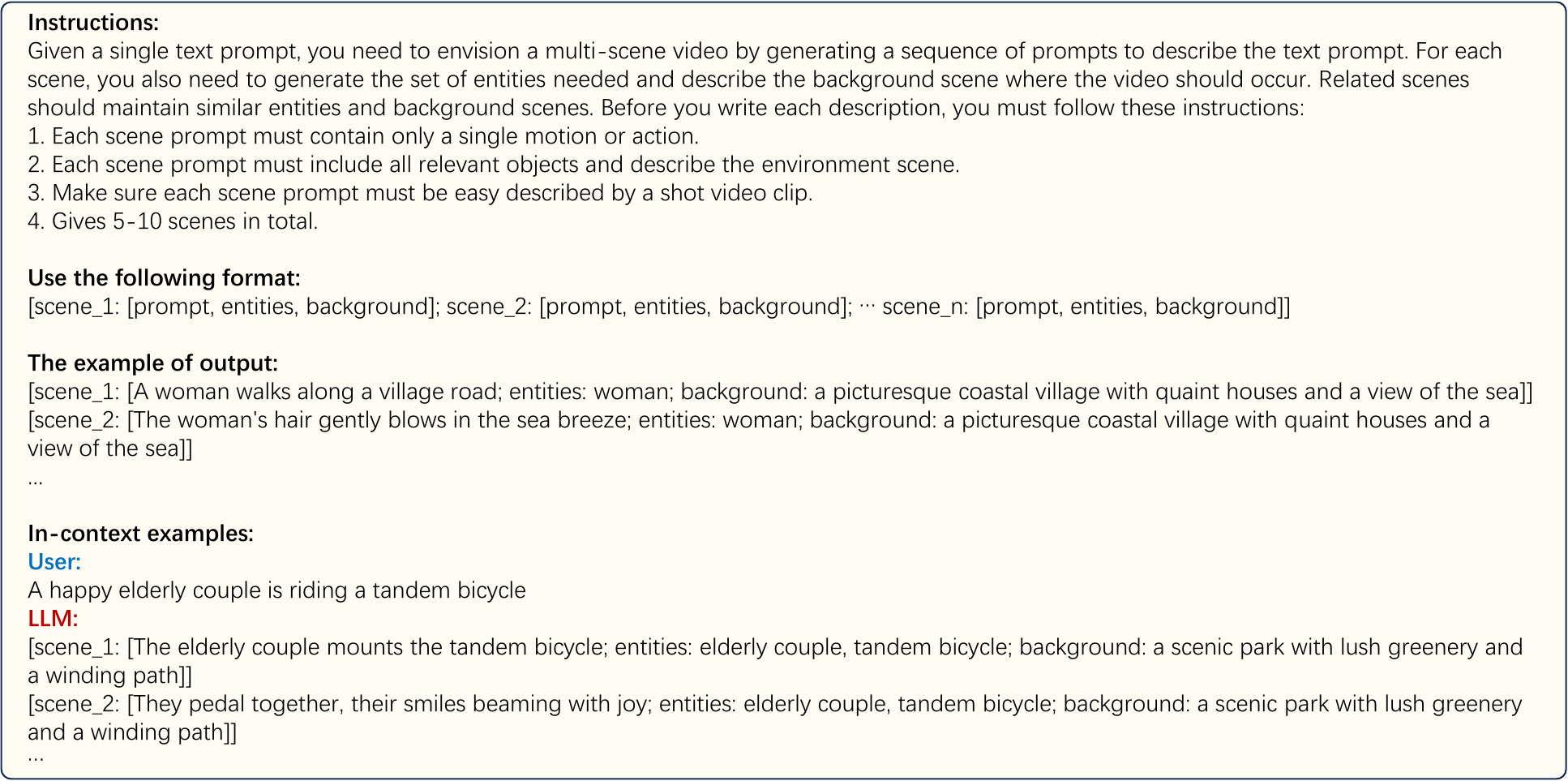}
	\vspace{-0.0in}
	\caption{The LLM instructions, output examples and in-context examples for generating scene prompts and common entities.}
	\label{fig:prompts-a}
\end{figure*}

\begin{figure*}[tbp]
	\centering
	\includegraphics[width=0.96\textwidth]{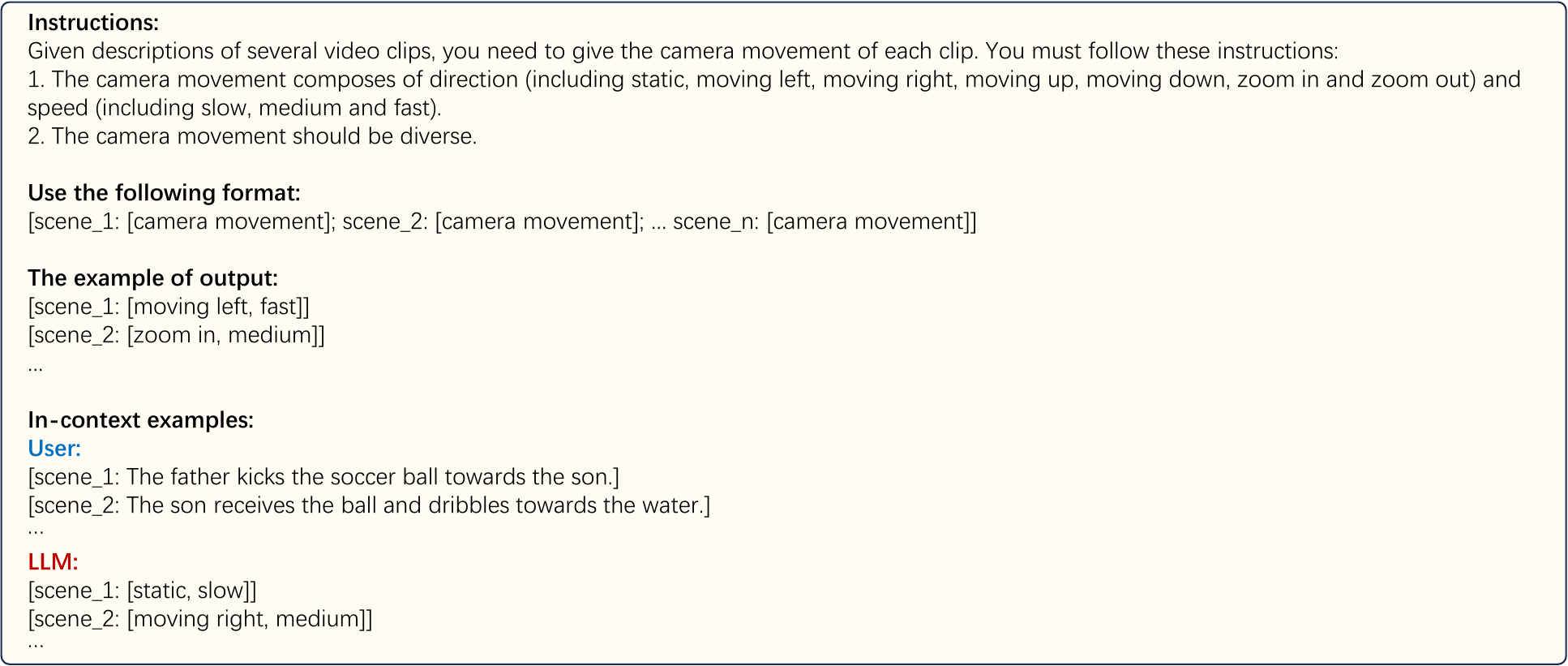}
	\vspace{-0.0in}
	\caption{The LLM instructions, output examples and in-context examples for generating camera movements.}
	\label{fig:prompts-b}
\end{figure*}

\begin{figure*}[tbp]
	\centering
	\includegraphics[width=0.96\textwidth]{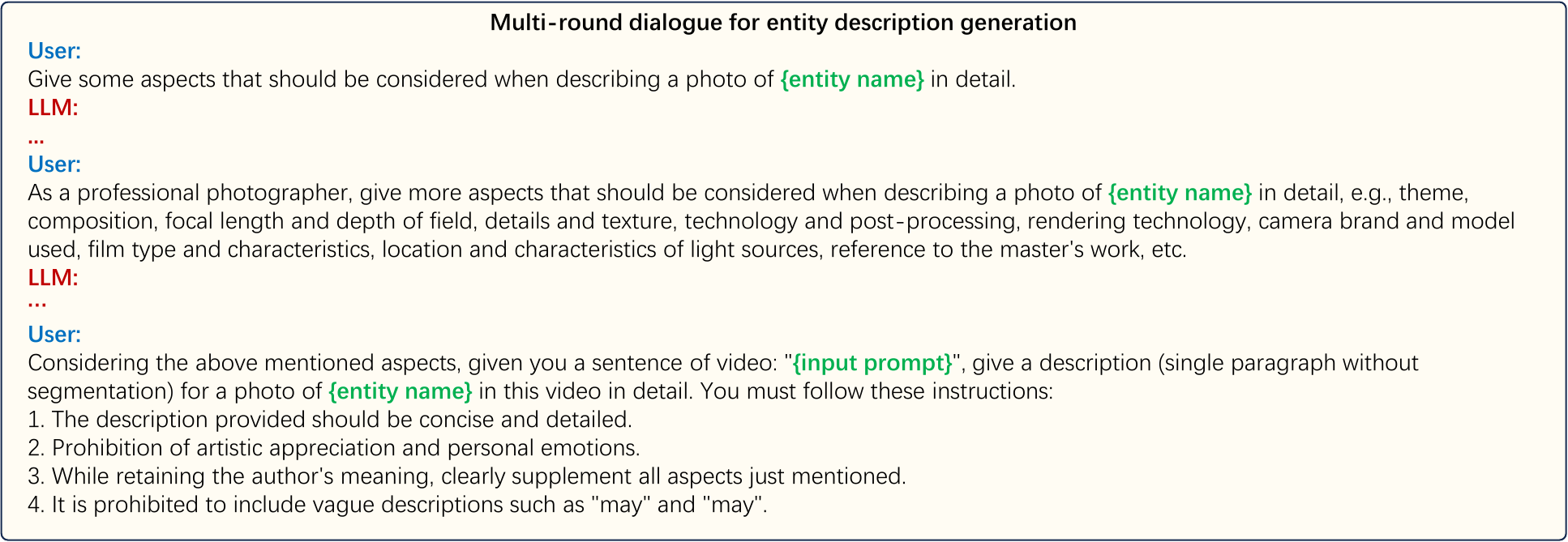}
	\vspace{-0.0in}
	\caption{The multi-round dialogue of LLM to achieve detailed entity description.}
	\label{fig:prompts-c}
\end{figure*}

The LLM instructions, output examples and in-context examples for video script and entity description generation are given in Figure~\ref{fig:prompts-a} and Figure~\ref{fig:prompts-b}, respectively. The multi-round dialogue for entity description generation is shown in Figure~\ref{fig:prompts-c}.

\section{Implementation details of VideoStudio-Img}
VideoStudio-Img is constructed on Stable Diffusion v2.1 model by incorporating the two additional cross-attention modules.
Table \ref{tab:img-param} details the structures of VideoStudio-Img.
We utilize the CLIP ViT-H/14 \cite{radford2021learning} as the text and visual encoder to extract text features from text prompt, and local image features from foreground/background reference image, respectively.
The sequence length $L_t$ of the text features is $77$ while the length $L_f$/$L_b$ of foreground/background image features is set as $256$.
The cross attention dimension $C_t$ and $C_f$/$C_b$ are set as the default number in each block of the original diffusion model.

\begin{table}[tbp]
	\centering
	\caption{Detailed hyper-parameters and structures of VideoStudio-Img.}
	\vspace{-0.1in}
	\resizebox{0.75\textwidth}{!}{
		\begin{tabular}{@{~}l@{~}c@{~}|@{~}l@{~}c@{~}}
			\toprule
			{Hyper-parameter} & {Value} & Hyper-parameter & Value\\ \midrule
			Base structure & SD v2.1 & Spatial transformer blocks & [1, 1, 1, 1] \\
			Latent shape & 4 $\times$ 64 $\times$ 64 & Image embed sequence & 256 \\
			Channels & 320 & Text CLIP & CLIP ViT-H/14 \\
			Layers per block & 2 & Parameterization & $\epsilon$ \\
			Channel multiplier &  [1, 2, 4, 4] & Diffusion steps & 1000 \\
			Attention resolutions & [64, 32, 16] & Noise schedule & Scaled Linear \\
			Head channels & [5, 10, 20] & $\beta_1$ & 0.00085 \\
			Number of heads & 64 & $\beta_T$ & 0.0120 \\
			CA embed dim & 1024 & Sampler & DDIM \\
			CA resolutions & [64, 32, 16] & Inference steps & 50 \\
			Autoencoders & AutoKL & GPU Type & A100-80G \\
			Image CLIP   & CLIP ViT-H/14 & GPU Number & 64 \\
			Learning rate    & $1\times 10^{-4}$ & Train steps & 20K \\
			Total batch size & 512 & $\#$ of UNet params & 915M  \\
			\bottomrule
		\end{tabular}
	}
	\label{tab:img-param}
	\vspace{-0.1in}
\end{table}

\begin{table}[tbp]
	\centering
	\caption{Detailed hyper-parameters and structures of VideoStudio-Vid.}
	\vspace{-0.1in}
	\resizebox{0.75\textwidth}{!}{
		\begin{tabular}{@{~}l@{~}c@{~}|@{~}l@{~}c@{~}}
			\toprule
			{Hyper-parameter} & {Value} & Hyper-parameter & Value\\ \midrule
			Base structure & SD-XL & Spatial transformer blocks & [0, 2, 10] \\
			Latent shape & 4 $\times$ 16 $\times$ 40 $\times$ 64 & Temporal transformer blocks & [0, 2, 10] \\
			Channels & 320 & Temporal SA head number & 64 \\
			Layers per block & 2 & Diffusion steps & 1000 \\
			Channel multiplier &  [1, 2, 4] & Noise schedule & Scaled Linear \\
			Attention resolutions & [32, 16] & $\beta_1$ & 0.00085 \\
			Head channels & [10, 20] & $\beta_T$ & 0.0120 \\
			Number of heads & 64 & Sampler & DDIM \\
			CA embed dim & 1280 & Inference steps & 70 \\
			CA resolutions & [32, 16] & $\eta$ & 1.0 \\
			Autoencoders & AutoKL & Guidance scale & 12.0 \\
			Image CLIP   & CLIP ViT-H/14 & GPU Type & A100-80G \\
			Parameterization & $\epsilon$ & GPU Number & 64 \\
			Learning rate & $3\times 10^{-6}$ & Train steps & 480K  \\
			Total batch size & 128 & $\#$ of 3D-UNet params & 4.7B  \\
			\bottomrule
		\end{tabular}
	}
	\label{tab:vid-param}
\end{table}

\section{Implementation details of VideoStudio-Vid}
We build the 3D UNet of VideoStudio-Vid by inserting temporal transformer and temporal convolution layers into 2D UNet of SD-XL.
Table \ref{tab:vid-param} details the hyper-parameters and structures of VideoStudio-Vid.
We employ the CLIP ViT-H/14 as the visual encoder to extract image features from scene-reference images.
To enhance the visual alignment between the scene-reference image and synthesized video, we concatenate the latent code of the scene-reference image with the noisy video latent code along temporal dimension as the input of 3D UNet.

\textbf{Action condition.} In the stage of model training, the VideoMAE \cite{tong2022vmae} fine-tuned on Kinetics-400 \cite{carreira2017quo} is leveraged as the action classifier to measure the action probability (i.e., indicator vector) $y_a$ of input videos.
A linear embedding is then learnt on the probability $y_a$ and further treated as a condition to adjust the video diffusion as demonstrated in the main paper.
In inference stage, we use the spaCy library to extract all action phrases from the input text prompt.
Next, the text features of the action phrases are obtained by using CLIP model, which are further exploited for cosine similarity computation with action vocabulary of Kinetics-400.
For each action phrase, we choose the action category with the max cosine similarity score.
If the cosine similarity is lower than $0.2$, the action category will be dropped.
After collecting all action categories and corresponding cosine similarity, we construct the action indicator vector $y_a\in [0,1]^{400}$ by assigning the normalized cosine similarity into the corresponding category index.

\begin{figure}[tbp]
	\centering
	\includegraphics[width=0.6\textwidth]{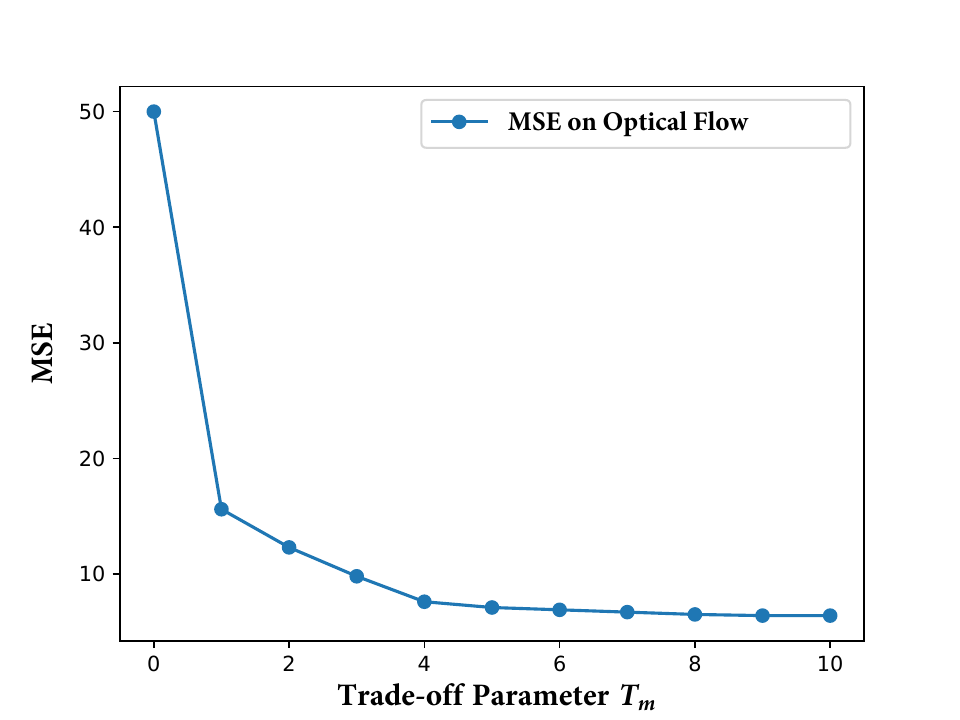}
	\vspace{-0.05in}
	\caption{The impact of trade-off parameter $T_m$ for camera movement.}
	\label{fig:tradeoff}
	\vspace{-0.05in}
\end{figure}

\textbf{Camera movement.} We control the camera movement of each scene video during the inference process of VideoStudio-Vid. Specifically, at inference timestep $t$, the noisy video $x_t=\alpha_t x_0 + \sigma_t \hat{\epsilon_t}$ is decomposed into the clean video $x_0$ with an estimated noise $\hat{\epsilon_t}=\epsilon_\theta\left(x_t, t\right)$ with fixed scheduling weights $\alpha_t$ and $\sigma_t$. The noisy video $x_t$ is transformed into a video $x_{t-1}$ with reduced noise:
\begin{equation}
	\small
	\begin{aligned}
		x_{t-1}=sampling(x_t, \hat{\epsilon_t}, t),
	\end{aligned}
	\label{eq:sampling-dpm}
\end{equation}
where $x_{T}$ represents the pure noise $\epsilon_T$. $sampling$ is the DDIM~\cite{song2022denoising} update strategy. After the first $T_m$ steps, we execute an adjustment to the noisy video $x_{(T_m-1)}$ to maintain the camera movement indicated by the video script as
\begin{equation}
	\small
	\begin{aligned}
		& \hat{x_0}=(x_{(T_m-1)} - \sigma_{(T_m-1)} \hat{\epsilon}_{T_m})/\alpha_{(T_m-1)}, \\
		& \overline{x_0}=0.5 \times \hat{x_0} + 0.5 \times warp(\hat{x_0}, flow), \\
		& \overline{x_{(T_m-1)}}=\alpha_{(T_m-1)} \overline{x_0} + \sigma_{(T_m-1)} \hat{\epsilon_{T_m}},
	\end{aligned}
	\label{eq:sampling-cm}
\end{equation}
where $warp(\hat{x_0}, flow)$ is to warp the frames in $\hat{x_0}$ based on the optical flow of required camera movement, and $\overline{x_{(T_m-1)}}$ is the modified noisy video. Such modification ensures that the resultant video clip maintains the same camera movement as we warp the intermediate frames. 
To analysis the impact of hyper-parameter $T_m$, we conduct the experiments on the MSR-VTT dataset and calculate the mean squared error (MSE) between the optical flow of generated and target videos with different $T_m$, as shown in Figure~\ref{fig:tradeoff}. 
In general, setting a small $T_m$ for early modification may not effectively control the camera movement, while a late modification may affect the visual quality of the generated videos. As indicated by the figure, $T_m$=5 provides a good trade-off empirically.

\begin{table}[t]
	\centering
	\caption{Performance comparison on ActivityNet Captions dataset.}
	\vspace{-0.05in}
	\setlength{\tabcolsep}{0.6cm}\resizebox{1.0\linewidth}{!}{
		\begin{tabular}{l|c|c|c|c|c|c}
			\toprule
			{Approach} & Ref & Training Data & Architecture & {FID} ($\downarrow$)  & {FVD} ($\downarrow$) & {Scene Consis.} ($\uparrow$) \\ \midrule
			ModelScopeT2V \cite{2023modelscope}    		    & 		    & LAION + WebVid-10M &  SD-2.1         & 18.1 & 980 & 46.0 \\
			VideoDirectorGPT \cite{lin2023videodirectorgpt} &          & LAION + WebVid-10M + GLIGEN &  SD-2.1         & 16.5 & 805 & 64.8 \\ \midrule
			\multirow{3}{*}{VideoStudio} & & LAION + WebVid-10M 		 &  SD-XL & 17.7 & 789 & 49.1 \\
			& & LAION + WebVid-10M + HD-VG &  SD-XL & 17.3 & 624 & 50.8 \\
			& \checkmark & LAION + WebVid-10M + HD-VG &  SD-XL  & \textbf{13.2} & \textbf{395} & \textbf{75.1}  \\
			\bottomrule
		\end{tabular}
	}
	\label{tab:scene.anet.sup}
	\vspace{-0.1in}
\end{table}

\section{Performance Contribution of VideoStudio}
To ablate the performance contribution of VideoStudio more transparent from different perspectives (e.g., with or without reference image, training data and architecture), we report the performances of different VideoStudio variants on ActivityNet Captions dataset in Table \ref{tab:scene.anet.sup}. The first variant is trained on LAION and WebVid-10M, which is the same as in ModelScopeT2V. The improvement over ModelScopeT2V in video quality (FID \& FVD) is due to the deeper Stable Diffusion (SD) backbone and the two-stage (T2I+I2V) framework. The second variant further utilizes HD-VG dataset for model training and leads to slightly better video quality. The full version of VideoStudio takes entity reference images into consideration and improves both video quality and cross-scene consistency.

\section{More Video Examples}
\begin{figure*}[tbp]
	\centering
	\includegraphics[width=0.98\textwidth]{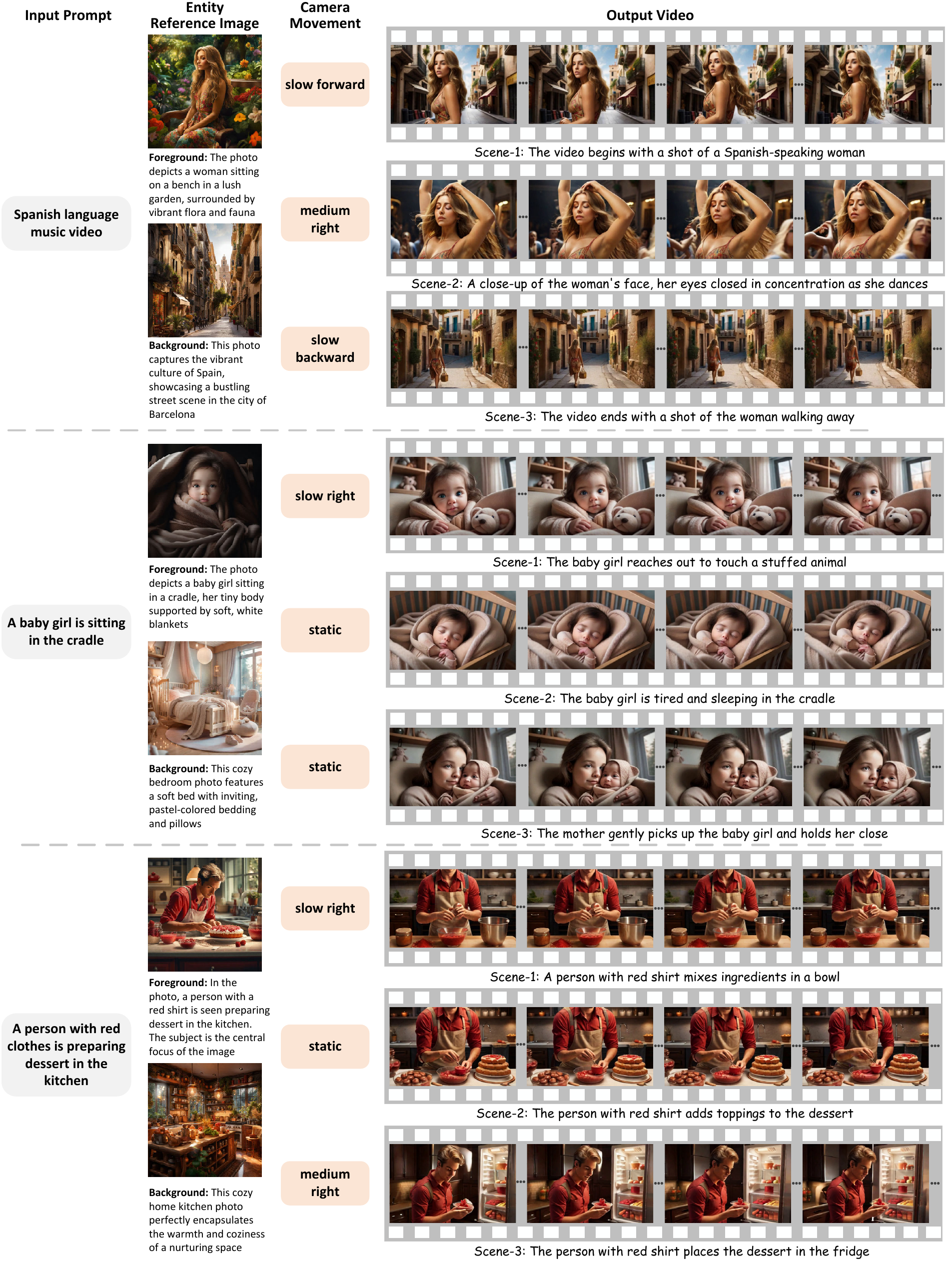}
	\vspace{-0.0in}
	\caption{Three examples of generated multi-scene videos by VideoStudio on MSR-VTT with single foreground reference image.}
	\label{fig:pipe.sup}
\end{figure*}

\begin{figure*}[tbp]
	\centering
	\includegraphics[width=0.99\textwidth]{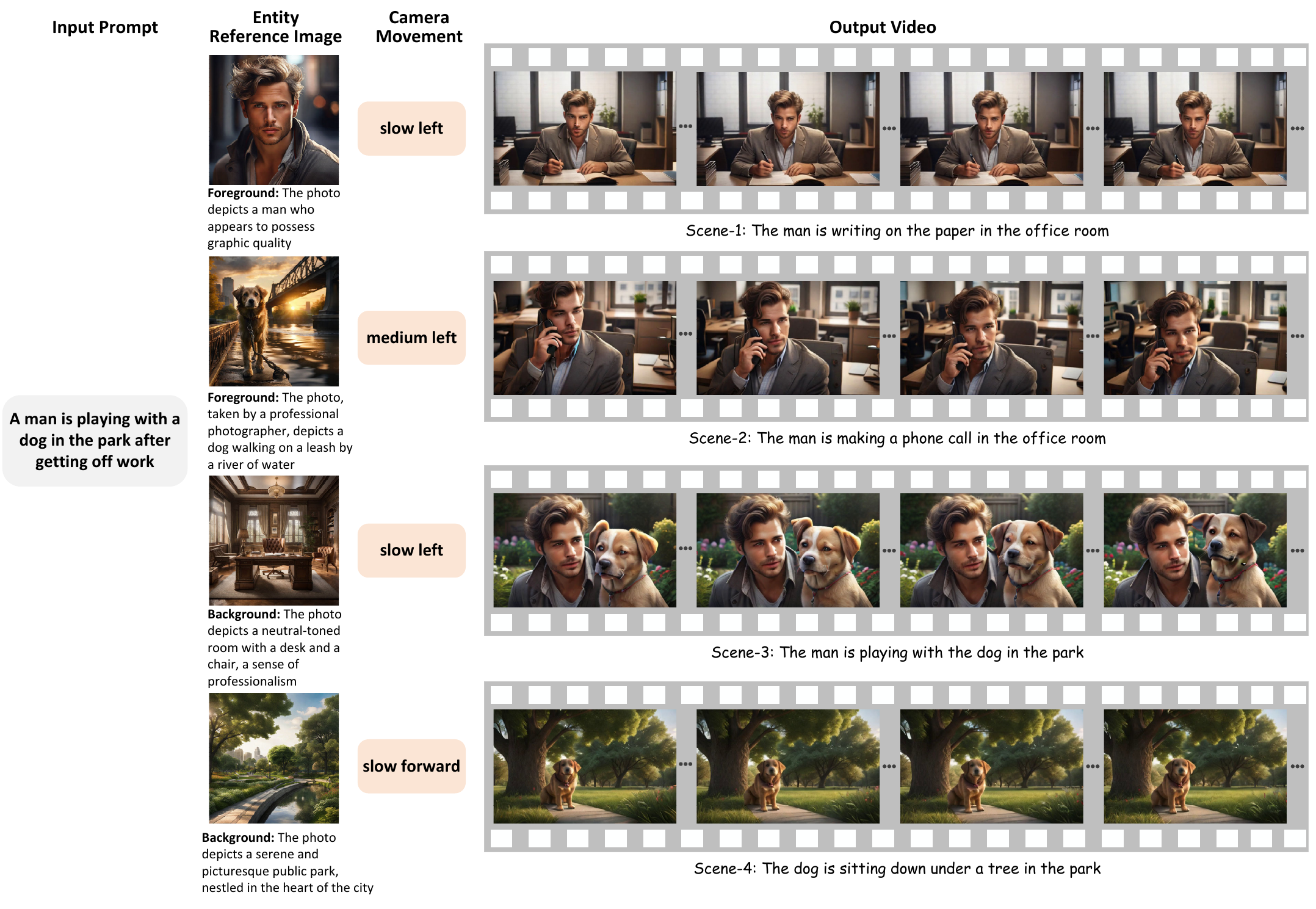}
	\vspace{-0.0in}
	\caption{One example of generated multi-scene videos by VideoStudio on MSR-VTT with multiple foreground reference images.}
	\label{fig:multi}
\end{figure*}

\begin{figure*}[tbp]
	\centering
	\includegraphics[width=0.95\textwidth]{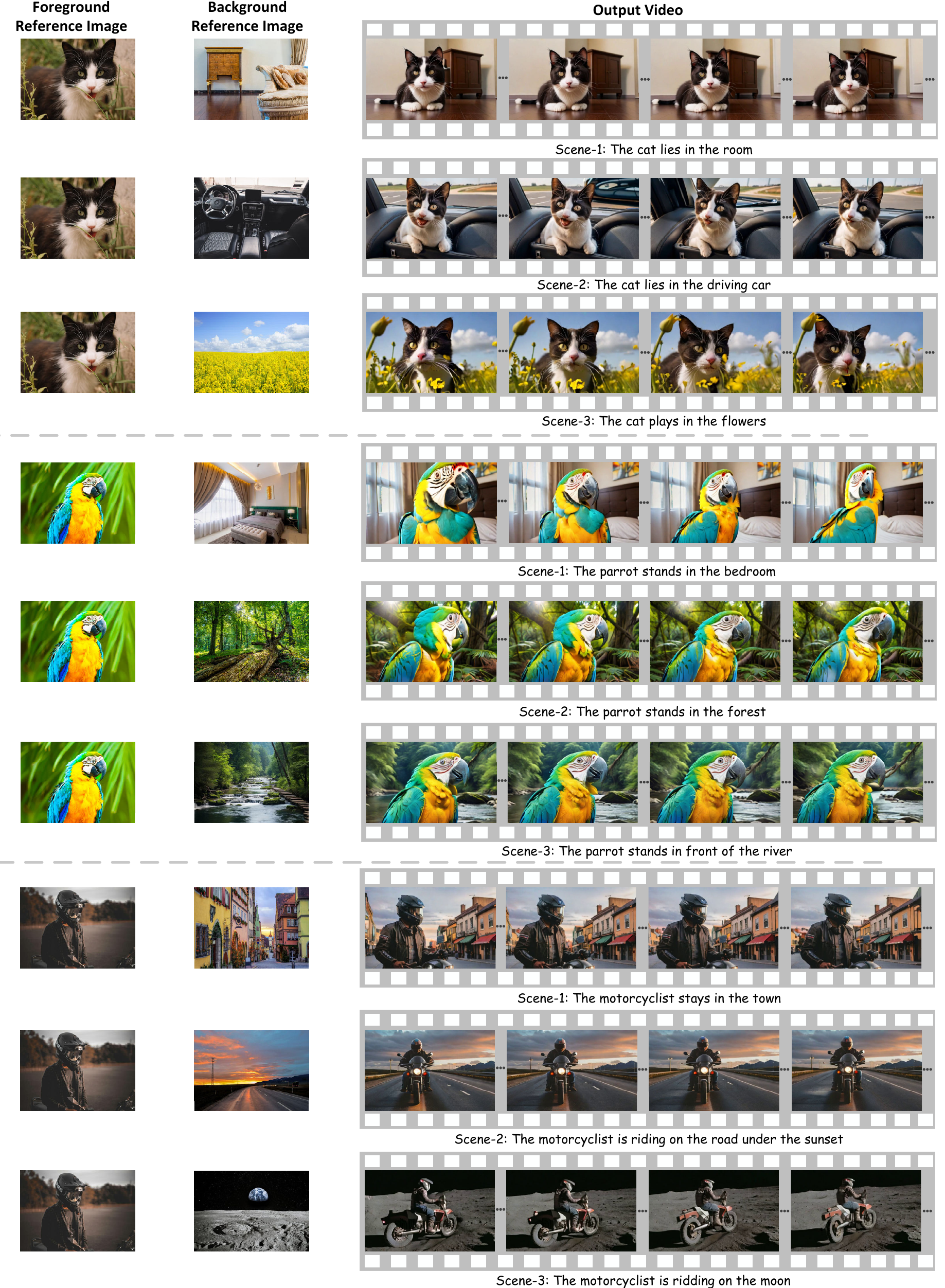}
	\vspace{-0.0in}
	\caption{Three example of generated multi-scene videos by our VideoStudio using the real images as entity reference images.}
	\label{fig:real.sup}
\end{figure*}

Here, we present more examples of multi-scene videos generated by VideoStudio on MSR-VTT in Figure~\ref{fig:pipe.sup} and Figure~\ref{fig:multi} with single foreground reference image and multiple foreground reference images, respectively. For each example, the input prompt, camera movement, foreground/background reference images and generated multi-scene video are given. Figure~\ref{fig:real.sup} further showcases three examples of generated multi-scene videos by VideoStudio using the real images as the entity reference images including foreground and background reference images. 

\section{Video Demo}
We have provided a video demo (VideoStudio.mp4) to illustrate the generated videos by VideoStudio in various scenarios. For more details, please refer to the offline project page (VideoStudio.html).

\clearpage

%% file: VideoStudio_camera.bbl
\begin{thebibliography}{10}
\providecommand{\url}[1]{\texttt{#1}}
\providecommand{\urlprefix}{URL }
\providecommand{\doi}[1]{https://doi.org/#1}

\bibitem{bain2021frozen}
Bain, M., Nagrani, A., Varol, G., Zisserman, A.: {Frozen in Time: A Joint Video
  and Image Encoder for End-to-End Retrieval}. In: ICCV (2021)

\bibitem{blattmann2023svd}
Blattmann, A., Dockhorn, T., Kulal, S., Mendelevitch, D., Kilian, M., Lorenz,
  D., Levi, Y., English, Z., Voleti, V., Letts, A., Jampani, V., Rombach, R.:
  {Stable Video Diffusion: Scaling Latent Video Diffusion Models to Large
  Datasets}. arXiv preprint arXiv:2311.15127  (2023)

\bibitem{blattmann2023align}
Blattmann, A., Rombach, R., Ling, H., Dockhorn, T., Kim, S.W., Fidler, S.,
  Kreis, K.: {Align your Latents: High-Resolution Video Synthesis with Latent
  Diffusion Models}. In: CVPR (2023)

\bibitem{carreira2017quo}
Carreira, J., Zisserman, A.: {Quo Vadis, Action Recognition? A New Model and
  The Kinetics Dataset}. In: CVPR (2017)

\bibitem{chen2023videocrafter1}
Chen, H., Xia, M., He, Y., Zhang, Y., Cun, X., Yang, S., Xing, J., Liu, Y.,
  Chen, Q., Wang, X., Weng, C., Shan, Y.: {VideoCrafter1: Open Diffusion Models
  for High-Quality Video Generation}. arXiv preprint arXiv:2310.19512  (2023)

\bibitem{chen2024sateco}
Chen, Z., Long, F., Qiu, Z., Yao, T., Zhou, W., Luo, J., Mei, T.: {Learning
  Spatial Adaptation and Temporal Coherence in Diffusion Models for Video
  Super-Resolution}. In: CVPR (2024)

\bibitem{dhariwal2021diffusion}
Dhariwal, P., Nichol, A.: {Diffusion Models Beat GANs on Image Synthesis}. In:
  NeurIPS (2021)

\bibitem{du2022glm}
Du, Z., Qian, Y., Liu, X., Ding, M., Qiu, J., Yang, Z., Tang, J.: {GLM: General
  Language Model Pretraining with Autoregressive Blank Infilling}. In: ACL
  (2022)

\bibitem{esser2023structure}
Esser, P., Chiu, J., Atighehchian, P., Granskog, J., Germanidis, A.: {Structure
  and Content-guided Video Synthesis with Diffusion Models}. In: ICCV (2023)

\bibitem{geyer2023tokenflow}
Geyer, M., Bar-Tal, O., Bagon, S., Dekel, T.: {TokenFlow: Consistent Diffusion
  Features for Consistent Video Editing}. arXiv preprint arXiv:2307.10373
  (2023)

\bibitem{guo2023animatediff}
Guo, Y., Yang, C., Rao, A., Wang, Y., Qiao, Y., Lin, D., Dai, B.: {AnimateDiff:
  Animate Your Personalized Text-to-Image Diffusion Models without Specific
  Tuning}. arXiv preprint arXiv:2307.04725  (2023)

\bibitem{he2022lvdm}
He, Y., Yang, T., Zhang, Y., Shan, Y., Chen, Q.: {Latent Video Diffusion Models
  for High-Fidelity Long Video Generation}. arXiv preprint arXiv:2211.13221
  (2022)

\bibitem{heusel2017gans}
Heusel, M., Ramsauer, H., Unterthiner, T., Nessler, B., Hochreiter, S.: {Gans
  Trained by a Two Time-Scale Update Rule Converge to a Local Nash
  Equilibrium}. In: NIPS (2017)

\bibitem{ho2022imagen}
Ho, J., Chan, W., Saharia, C., Whang, J., Gao, R., Gritsenko, A., Kingma, D.P.,
  Poole, B., Norouzi, M., Fleet, D.J., Salimans, T.: {Imagen Video: High
  Definition Video Generation with Diffusion Models}. arXiv preprint
  arXiv:2210.02303  (2022)

\bibitem{ho2020denoising}
Ho, J., Jain, A., Abbeel, P.: {Denoising Diffusion Probabilistic Models}. In:
  NeurIPS (2020)

\bibitem{ho2022classifier}
Ho, J., Salimans, T.: {Classifier-Free Diffusion Guidance}. arXiv preprint
  arXiv:2207.12598  (2022)

\bibitem{2022cogvideo}
Hong, W., Ding, M., Zheng, W., Liu, X., Tang, J.: {CogVideo: Large-Scale
  Pretraining for Text-to-Video Generation via Transformers}. In: ICLR (2023)

\bibitem{hu2023videocontrolnet}
Hu, Z., Xu, D.: {VideoControlNet: A Motion-Guided Video-to-Video Translation
  Framework by Using Diffusion Model with ControlNet}. arXiv preprint
  arXiv:2307.14073  (2023)

\bibitem{huang2023vbench}
Huang, Z., He, Y., Yu, J., Zhang, F., Si, C., Jiang, Y., Zhang, Y., Wu, T.,
  Jin, Q., Chanpaisit, N., Wang, Y., Chen, X., Wang, L., Lin, D., Qiao, Y.,
  Liu, Z.: {VBench: Comprehensive Benchmark Suite for Video Generative Models}.
  arXiv preprint arXiv:2311.17982  (2023)

\bibitem{khachatryan2023text2videozero}
Khachatryan, L., Movsisyan, A., Tadevosyan, V., Henschel, R., Wang, Z.,
  Navasardyan, S., Shi, H.: {Text2Video-Zero: Text-to-Image Diffusion Models
  are Zero-Shot Video Generators}. In: ICCV (2023)

\bibitem{kim2017deepstory}
Kim, K.M., Heo, M.O., Choi, S.H., Zhang, B.T.: {DeepStory: Video Story QA by
  Deep Embedded Memory Networks}. In: IJCAI (2017)

\bibitem{krishna2017dense}
Krishna, R., Hata, K., Ren, F., Fei-Fei, L., Niebles, J.C.: {Dense-Captioning
  Events in Videos}. In: ICCV (2017)

\bibitem{li2022CoT}
Li, Y., Yao, T., Pan, Y., Mei, T.: {Contextual Transformer Networks for Visual
  Recognition}. IEEE Trans. on PAMI  (2022)

\bibitem{li2019storygan}
Li, Y., Gan, Z., Shen, Y., Liu, J., Cheng, Y., Wu, Y., Carin, L., Carlson, D.,
  Gao, J.: {StoryGAN: A Sequential Conditional GAN for Story Visualization}.
  In: CVPR (2019)

\bibitem{liang2022nuwa}
Liang, J., Wu, C., Hu, X., Gan, Z., Wang, J., Wang, L., Liu, Z., Fang, Y.,
  Duan, N.: {NUWA-Infinity: Autoregressive over Autoregressive Generation for
  Infinite Visual Synthesis}. In: NeurIPS (2022)

\bibitem{lin2023videodirectorgpt}
Lin, H., Zala, A., Cho, J., Bansal, M.: {VideoDirectorGPT: Consistent
  Multi-Scene Video Generation via LLM-Guided Planning}. arXiv preprint
  arXiv:2309.15091  (2023)

\bibitem{liu2023grounding}
Liu, S., Zeng, Z., Ren, T., Li, F., Zhang, H., Yang, J., Li, C., Yang, J., Su,
  H., Zhu, J., et~al.: {Grounding DINO: Marrying Dino with Grounded
  Pre-Training for Open-Set Object Detection}. arXiv preprint arXiv:2303.05499
  (2023)

\bibitem{long2022sifa}
Long, F., Qiu, Z., Pan, Y., Yao, T., Luo, J., Mei, T.: {Stand-Alone Inter-Frame
  Attention in Video Models}. In: CVPR (2022)

\bibitem{long2022dynamic}
Long, F., Qiu, Z., Pan, Y., Yao, T., Ngo, C.W., Mei, T.: {Dynamic Temporal
  Filtering in Video Models}. In: ECCV (2022)

\bibitem{long2019gaussian}
Long, F., Yao, T., Qiu, Z., Tian, X., Luo, J., Mei, T.: {Gaussian Temporal
  Awareness Networks for Action Localization}. In: CVPR (2019)

\bibitem{long2023BCN}
Long, F., Yao, T., Qiu, Z., Tian, X., Luo, J., Mei, T.: {Bi-calibration
  Networks for Weakly-Supervised Video Representation Learning}. IJCV  (2023)

\bibitem{lu2022dpmsolver}
Lu, C., Zhou, Y., Bao, F., Chen, J., Li, C., Zhu, J.: {{DPM}-Solver: A Fast
  {ODE} Solver for Diffusion Probabilistic Model Sampling in Around 10 Steps}.
  In: NeurIPS (2022)

\bibitem{lu2023dpmsolver}
Lu, C., Zhou, Y., Bao, F., Chen, J., Li, C., Zhu, J.: {{DPM}-Solver++: Fast
  Solver for Guided Sampling of Diffusion Probabilistic Models}. arXiv preprint
  arXiv:2211.01095  (2023)

\bibitem{luo2023videofusion}
Luo, Z., Chen, D., Zhang, Y., Huang, Y., Wang, L., Shen, Y., Zhao, D., Zhou,
  J., Tan, T.: {VideoFusion: Decomposed Diffusion Models for High-Quality Video
  Generation}. In: CVPR (2023)

\bibitem{mou2023t2iadapter}
Mou, C., Wang, X., Xie, L., Wu, Y., Zhang, J., Qi, Z., Shan, Y., Qie, X.:
  {T2I-Adapter: Learning Adapters to Dig out More Controllable Ability for
  Text-to-Image Diffusion Models}. arXiv preprint arXiv:2302.08453  (2023)

\bibitem{nichol2021improved}
Nichol, A., Dhariwal, P.: {Improved Denoising Diffusion Probabilistic Models}.
  In: ICML (2021)

\bibitem{nichol2022glide}
Nichol, A., Dhariwal, P., Ramesh, A., Shyam, P., Mishkin, P., McGrew, B.,
  Sutskever, I., Chen, M.: {GLIDE: Towards Photorealistic Image Generation and
  Editing with Text-Guided Diffusion Models}. In: ICML (2022)

\bibitem{openai2023gpt4}
OpenAI: {GPT-4 Technical Report} (2023)

\bibitem{ouyang2023codef}
Ouyang, H., Wang, Q., Xiao, Y., Bai, Q., Zhang, J., Zheng, K., Zhou, X., Chen,
  Q., Shen, Y.: {CoDeF: Content Deformation Fields for Temporally Consistent
  Video Processing}. arXiv preprint arXiv:2308.07926  (2023)

\bibitem{qi2023fatezero}
Qi, C., Cun, X., Zhang, Y., Lei, C., Wang, X., Shan, Y., Chen, Q.: {FateZero:
  Fusing Attentions for Zero-shot Text-based Video Editing}. In: ICCV (2023)

\bibitem{Qin_2020_PR}
Qin, X., Zhang, Z., Huang, C., Dehghan, M., Zaiane, O., Jagersand, M.: {U2-Net:
  Going Deeper with Nested U-Structure for Salient Object Detection}. Pattern
  Recognition  (2020)

\bibitem{radford2021learning}
Radford, A., Kim, J.W., Hallacy, C., Ramesh, A., Goh, G., Agarwal, S., Sastry,
  G., Askell, A., Mishkin, P., Clark, J., Krueger, G., Sutskever, I.: {Learning
  Transferable Visual Models from Natural Language Supervision}. In: ICML
  (2021)

\bibitem{ramesh2022hierarchical}
Ramesh, A., Dhariwal, P., Nichol, A., Chu, C., Chen, M.: {Hierarchical
  Text-Conditional Image Generation with CLIP Latents}. arXiv preprint
  arXiv:2204.06125  (2022)

\bibitem{rombach2022high}
Rombach, R., Blattmann, A., Lorenz, D., Esser, P., Ommer, B.: {High-Resolution
  Image Synthesis with Latent Diffusion Models}. In: CVPR (2022)

\bibitem{schuhmann2022laion}
Schuhmann, C., Beaumont, R., Vencu, R., Gordon, C., Wightman, R., Cherti, M.,
  Coombes, T., Katta, A., Mullis, C., Wortsman, M., et~al.: {Laion-5B: An Open
  Large-Scale Dataset for Training Next Generation Image-Text Models}. In:
  NeurIPS (2022)

\bibitem{shin2023editavideo}
Shin, C., Kim, H., Lee, C.H., gil Lee, S., Yoon, S.: {Edit-A-Video: Single
  Video Editing with Object-Aware Consistency}. arXiv preprint arXiv:2303.07945
   (2023)

\bibitem{singer2022make}
Singer, U., Polyak, A., Hayes, T., Yin, X., An, J., Zhang, S., Hu, Q., Yang,
  H., Ashual, O., Gafni, O., Parikh, D., Gupta, S., Taigman, Y.: {Make-a-video:
  Text-to-Video Generation without Text-Video Data}. arXiv preprint
  arXiv:2209.14792  (2022)

\bibitem{sohl2015deep}
Sohl-Dickstein, J., Weiss, E.A., Maheswaranathan, N., Ganguli, S.: {Deep
  Unsupervised Learning using Nonequilibrium Thermodynamics}. In: ICML (2015)

\bibitem{song2022denoising}
Song, J., Meng, C., Ermon, S.: {Denoising Diffusion Implicit Models}. In: ICLR
  (2021)

\bibitem{song2019generative}
Song, Y., Ermon, S.: {Generative Modeling by Estimating Gradients of the Data
  Distribution}. In: NeurIPS (2019)

\bibitem{unterthiner2019fvd}
Unterthiner, T., van Steenkiste, S., Kurach, K., Marinier, R., Michalski, M.,
  Gelly, S.: {FVD: A New Metric for Video Generation}. In: ICLR Workshop (2019)

\bibitem{villegas2022phenaki}
Villegas, R., Babaeizadeh, M., Kindermans, P.J., Moraldo, H., Zhang, H.,
  Saffar, M.T., Castro, S., Kunze, J., Erhan, D.: {Phenaki: Variable Length
  Video Generation from Open Domain Textual Description}. In: ICLR (2023)

\bibitem{voleti2022mcvd}
Voleti, V., Jolicoeur-Martineau, A., Pal, C.: {MCVD-Masked Conditional Video
  Diffusion for Prediction, Generation, and Interpolation}. In: NeurIPS (2022)

\bibitem{wang2023genlvideo}
Wang, F.Y., Chen, W., Song, G., Ye, H.J., Liu, Y., Li, H.: {Gen-L-Video:
  Multi-Text to Long Video Generation via Temporal Co-Denoising}. arXiv
  preprint arXiv:2305.18264  (2023)

\bibitem{2023modelscope}
Wang, J., Yuan, H., Chen, D., Zhang, Y., Wang, X., Zhang, S.: {ModelScope
  Text-to-Video Technical Report}. arXiv preprint arXiv:2308.06571  (2023)

\bibitem{wang2023videofactory}
Wang, W., Yang, H., Tuo, Z., He, H., Zhu, J., Fu, J., Liu, J.: {VideoFactory:
  Swap Attention in Spatiotemporal Diffusions for Text-to-Video Generation}.
  arXiv preprint arXiv:2305.10874  (2023)

\bibitem{wang2023videocomposer}
Wang, X., Yuan, H., Zhang, S., Chen, D., Wang, J., Zhang, Y., Shen, Y., Zhao,
  D., Zhou, J.: {VideoComposer: Compositional Video Synthesis with Motion
  Controllability}. In: NeurIPS (2023)

\bibitem{wu2021godiva}
Wu, C., Huang, L., Zhang, Q., Li, B., Ji, L., Yang, F., Sapiro, G., Duan, N.:
  {GODIVA: Generating Open-Domain Videos from Natural Descriptions}. arXiv
  preprint arXiv:2104.14806  (2021)

\bibitem{wu2023tuneavideo}
Wu, J.Z., Ge, Y., Wang, X., Lei, S.W., Gu, Y., Shi, Y., Hsu, W., Shan, Y., Qie,
  X., Shou, M.Z.: {Tune-A-Video: One-Shot Tuning of Image Diffusion Models for
  Text-to-Video Generation}. In: ICCV (2023)

\bibitem{xing2023dynamicrafter}
Xing, J., Xia, M., Zhang, Y., Chen, H., Yu, W., Liu, H., Wang, X., Wong, T.T.,
  Shan, Y.: {DynamiCrafter: Animating Open-domain Images with Video Diffusion
  Priors}. arXiv preprint arXiv:2310.12190  (2023)

\bibitem{xu2016msr}
Xu, J., Mei, T., Yao, T., Rui, Y.: {MSR-VTT: A Large Video Description Dataset
  for Bridging Video and Language}. In: CVPR (2016)

\bibitem{yao2023dual}
Yao, T., Li, Y., Pan, Y., Wang, Y., Zhang, X.P., Mei, T.: {Dual Vision
  Transformer}. IEEE Trans. on PAMI  (2023)

\bibitem{yao2022wave}
Yao, T., Pan, Y., Li, Y., Ngo, C.W., Mei, T.: {Wave-ViT: Unifying Wavelet and
  Transformers for Visual Representation Learning}. In: ECCV (2022)

\bibitem{ye2023ipadapter}
Ye, H., Zhang, J., Liu, S., Han, X., Yang, W.: {IP-Adapter: Text Compatible
  Image Prompt Adapter for Text-to-Image Diffusion Models}. arXiv preprint
  arXiv:2308.06721  (2023)

\bibitem{yin2023dragnuwa}
Yin, S., Wu, C., Liang, J., Shi, J., Li, H., Ming, G., Duan, N.: {Dragnuwa:
  Fine-grained Control in Video Generation by Integrating Text, Image, and
  Trajectory}. arXiv preprint arXiv:2308.08089  (2023)

\bibitem{yin2023nuwa}
Yin, S., Wu, C., Yang, H., Wang, J., Wang, X., Ni, M., Yang, Z., Li, L., Liu,
  S., Yang, F., Fu, J., Ming, G., Wang, L., Liu, Z., Li, H., Duan, N.:
  {NUWA-XL: Diffusion over Diffusion for eXtremely Long Video Generation}.
  arXiv preprint arXiv:2303.12346  (2023)

\bibitem{zeng2022glm}
Zeng, A., Liu, X., Du, Z., Wang, Z., Lai, H., Ding, M., Yang, Z., Xu, Y.,
  Zheng, W., Xia, X., et~al.: {GLM-130B: An Open Bilingual Pre-Trained Model}.
  arXiv preprint arXiv:2210.02414  (2022)

\bibitem{zhang2023adding}
Zhang, L., Rao, A., Agrawala, M.: {Adding Conditional Control to Text-to-Image
  Diffusion Models}. In: ICCV (2023)

\bibitem{zhang2023i2vgen}
Zhang, S., Wang, J., Zhang, Y., Zhao, K., Yuan, H., Qin, Z., Wang, X., Zhao,
  D., Zhou, J.: {I2VGen-XL: High-Quality Image-to-Video Synthesis via Cascaded
  Diffusion Models}. arXiv preprint arXiv:2311.04145  (2023)

\bibitem{zhang2024trip}
Zhang, Z., Long, F., Pan, Y., Qiu, Z., Yao, T., Cao, Y., Mei, T.: {TRIP:
  Temporal Residual Learning with Image Noise Prior for Image-to-Video
  Diffusion Models}. In: CVPR (2024)

\bibitem{2023magicvideo}
Zhou, D., Wang, W., Yan, H., Lv, W., Zhu, Y., Feng, J.: {Magicvideo: Efficient
  Video Generation with Latent Diffusion Models}. arXiv preprint
  arXiv:2211.11018  (2022)

\end{thebibliography}
